%% file: main.tex
\documentclass[letterpaper]{article} 
\usepackage[preprint]{scalehp}
\usepackage[hyphens]{url}  
\usepackage{graphicx} 
\usepackage{natbib}  
\usepackage{caption} 
\usepackage{amsmath}
\usepackage{booktabs}
\usepackage{array}
\usepackage{multirow}
\graphicspath{{Figures/ScaleHP/}{Figures/ScaleHP/Comparison/pics/}%
{Figures/ScaleHP_Supple/optimized_freihand/}%
{Figures/ScaleHP_Supple/optimized_wild/}{Figures/ScaleHP_Supple/}}
\title{ScaleHP: Scale-Mediated Optimization of Coupled Errors for Metric-Space Hand Pose Estimation}

\author{
    Ruitao Jing\textsuperscript{\rm 1,\rm 3,\rm 4}\equalcontrib\thanks{This work was done when Ruitao Jing was an intern at Visincept and IDEA Research.},
    Xingyu Chen\textsuperscript{\rm 2}\equalcontrib,
    Hongyang Li\textsuperscript{\rm 4,\rm 5},\\
    Qing Jiang\textsuperscript{\rm 4,\rm 5},
    Yukai Shi\textsuperscript{\rm 1,\rm 4},
    Lei Zhang\textsuperscript{\rm 3,\rm 4,\rm 5}\corresponding
}
\affiliations{
    \textsuperscript{\rm 1}Tsinghua University\quad
    \textsuperscript{\rm 2}Zhongguancun Academy\quad
    \textsuperscript{\rm 3}Visincept\\
    \textsuperscript{\rm 4}International Digital Economy Academy (IDEA Research)\quad
    \textsuperscript{\rm 5}South China University of Technology\\
    \url{https://laiang8086.github.io/scalehp}
}

\begin{document}

\maketitle

\begin{abstract}
In this paper, we present ScaleHP, a unified framework that explicitly represents
per-instance metric scale to resolve the coupled errors in calibrated camera-space
hand pose estimation. Under the common root-relative-to-global paradigm,
camera-space accuracy depends jointly on relative geometry, root localization,
and their scale-dependent composition. ScaleHP treats scale as the shared
interface among these terms rather than optimizing them in isolation. Its
metric-aware 2D--3D Transformer decoder introduces a dedicated scale token that
exchanges information with sparse, semantically indexed 2D/3D joint queries,
allowing global metric reasoning and local joint geometry to mutually refine one
another. A parameter-free calibrated module then recovers root translation from
the predicted unit-scale joints, metric scale, and known camera intrinsics.
ScaleHP demonstrates state-of-the-art CS-MPJPE on FreiHand (35.8\,mm) and,
under benchmark-specific training protocols, state-of-the-art PA-MPJPE on
FreiHand, DexYCB, and HO3Dv3 (5.0/4.6/5.9\,mm), showing that improved global
metric localization is attained together with improved local articulation.
Ablations further show that scale--pose interaction improves both camera-space
accuracy and relative depth geometry.
\end{abstract}

\section{Introduction}
\label{sec:intro}
Hand pose estimation (HPE) is a basic perception primitive for interaction,
manipulation, and teleoperation. In VR/AR, the hand must be placed consistently
with the surrounding scene; in embodied AI and robotic teleoperation, a
controller needs the hand's physical location and dimensions rather than only
its articulation
\citep{DBLP:journals/access/KarimKLANA25,DBLP:journals/thri/NostadtACB20,
DBLP:journals/firai/ToetKKE20}. These applications motivate metric camera-space
hand pose, where the output must preserve both the articulated configuration
and the hand's absolute position and size in the calibrated camera coordinate
system.

To obtain camera-space pose, a common paradigm first predicts root-relative
hand geometry and then places it globally. Writing
$\mathbf{Q}^{\mathrm{cam}}=s(\mathbf{J}^{\mathrm{rel}}+
\mathbf{t}^{\mathrm{norm}})$ exposes three coupled error sources: (a)
root-relative geometry in $\mathbf{J}^{\mathrm{rel}}$, (b) root localization
in $\mathbf{t}^{\mathrm{norm}}$, and (c) the cascade produced when the metric
scale $s$ multiplies the composed term. Existing methods address these terms with
different compromises. CMR \citep{DBLP:conf/cvpr/ChenLMCWCGWZ21} follows the
two-step recipe most explicitly: it predicts root-relative geometry and
registers the root to image evidence, leaving positioning decoupled from the
representation that produced the geometry. HandOS \citep{chen2025handos} brings
2D/3D queries and a translation variable into a one-stage decoder, but does not
organize translation as an independently supervised metric-distance factor;
reprojection alone cannot resolve the absolute metric gauge. HandDGP
\citep{valassakis2024handdgp} reconnects the stages with differentiable global
positioning, yet its analysis reports a trade-off between global positioning and
relative-geometry accuracy. NVF \citep{Huang_2023_CVPR} bypasses the factorization
with dense camera-frustum voting, using a substantially different sampling-based
formulation.
Overall, most prior systems primarily reduce error source~(a), namely
root-relative geometry. Error source~(b) is commonly delegated to a learned
translation or an external depth cue, which remains entangled with focal length
and camera distance. Error source~(c), the term most capable of widening
camera-space gaps, is either left implicit or improved only at the expense of
one of the first two terms. A related tension appears in metric-depth prediction: ScaleDepth
\citep{song2024scaledepth} and MoGe-2 \citep{wang2025moge} deliberately separate global metric quantities from local geometry,
and their designs make clear that the two predictions can
compete. This raises the central question of our work: can
the three error sources be optimized together, with global
metric reasoning and local hand geometry improving rather
than undermining one another?

\begin{figure*}[t]
    \centering
    \includegraphics[width=0.85\linewidth]{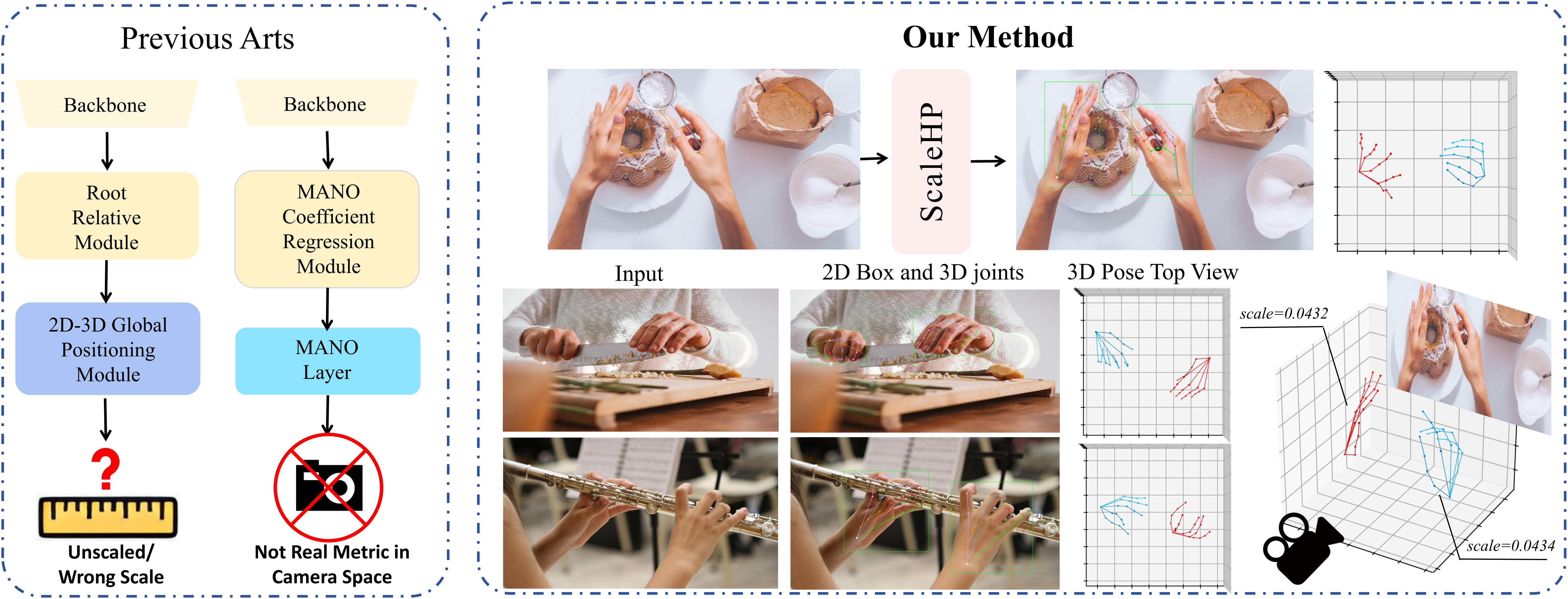}
    \caption{ScaleHP places metric scale at the center of calibrated camera-space
hand pose estimation. Unlike prior approaches that leave scale implicit, supply
it externally, or decouple it from pose learning, ScaleHP predicts per-instance
metric scale jointly with unit-scale root-relative geometry and uses analytically
recovered normalized translation to reconstruct the hand in camera space.}
    \label{fig:teaser}
\end{figure*}

We argue that the missing interface is an explicit instance scale. Under
calibrated perspective projection, the same 2D projections can arise from different hand
sizes paired with different depths. Without an explicit scale variable, existing
camera-space formulations---including MANO-based parametric pipelines and other
joint or mesh regressors---often misestimate metric scale during global
positioning, or incorrectly compensate for that error in the recovered root
translation. The absence of a separately supervised scale representation then
makes this coupling hard to diagnose. Metric-depth methods instead estimate dense scene geometry and
global scale from image-wise representations. Hand pose does not necessarily require a dense
scene representation: an individual hand pose can be described by sparse, semantically
indexed, and geometrically constrained 2D layouts and root-relative 3D joints.
These representations can provide scale evidence without a dense point map or an
external monocular-depth branch. Scale is therefore a natural bridge between
local articulation and global camera-space coordinates, while remaining a
learned statistical prior under metric supervision rather than a claim that
monocular projection uniquely identifies hand size.

\begin{figure*}[t]
    \centering
    \includegraphics[width=0.90\linewidth]{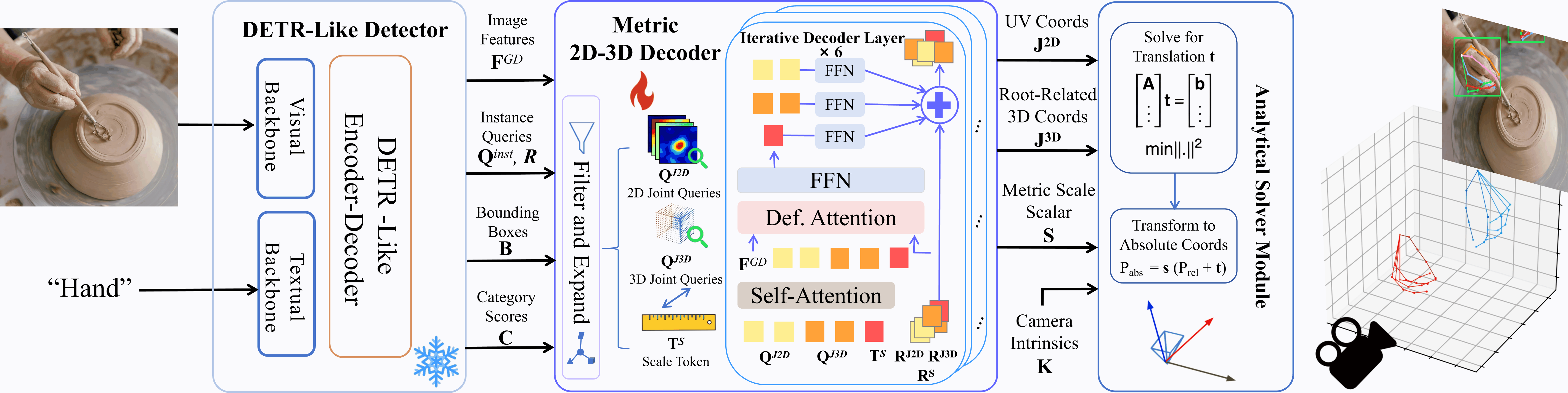}
    \caption{Overview of ScaleHP. A frozen detector supplies image and
instance features. The jointly trained decoder predicts 2D joints, unit-scale
root-relative 3D joints, and a per-instance metric scale through interacting
joint and scale tokens. A parameter-free calibrated solver then recovers
normalized translation and composes metric camera-space joints.}
\label{fig:overview}
\end{figure*}

Based on this insight, we present ScaleHP, a unified framework for calibrated
metric-space HPE (Fig.~\ref{fig:overview}). Its metric-aware 2D--3D Transformer
decoder introduces a hand-specific scale token that exchanges information with
sparse 2D/3D joint queries. Joint queries provide localized scale evidence, while
scale supervision reaches later joint-query layers instead of acting only as
post-hoc calibration. The decoder predicts 2D joints, unit-scale root-relative
3D joints, and instance metric scale from frozen-detector features. Given these
outputs and known intrinsics, a parameter-free least-squares module recovers
normalized root translation and composes metric camera-space joints. ScaleHP
thereby coordinates (a), (b), and their scale-dependent interaction (c) without
dense scene-depth prediction or iterative registration.

The contributions of this work are:
\begin{itemize}
    \item \textbf{Scale-mediated camera-space formulation.} We expose relative
geometry, root localization, and their scale-dependent composition as the three
coupled sources of camera-space error, and identify per-instance metric scale as
the interface between local hand structure and global coordinates.
    \item \textbf{Query-coupled ScaleHP.} We present a unified framework for
metric-space HPE whose scale token interacts with sparse semantic 2D/3D joint
queries, allowing global scale supervision and local joint geometry to mutually
refine one another. By jointly optimizing the learned relative-geometry and
scale factors while calibrated positioning resolves root localization under
known intrinsics, ScaleHP coordinates all three error sources in one pipeline.
    \item \textbf{State-of-the-art metric and relative accuracy.} ScaleHP
obtains the best listed FreiHand CS-MPJPE of 35.8\,mm. Under benchmark-specific
training protocols, it also achieves state-of-the-art PA-MPJPE of 5.0, 4.6, and
5.9\,mm on FreiHand, DexYCB, and HO3Dv3, respectively, showing that global
localization and local articulation improve together.
\end{itemize}

\section{Related Works}

\noindent\textbf{Camera-space HPE.}
Camera-space HPE must recover absolute placement and physical scale in addition
to articulation. Iqbal et al. \citep{iqbal2018handpose} lift a normalized pose
with a supplied or mean bone length. CMR and related registration-based systems
recover global position from 2D evidence and root-relative geometry
\citep{DBLP:conf/cvpr/ChenLMCWCGWZ21,
DBLP:conf/cvpr/ChenLDZMXZG22,DBLP:conf/cvpr/ParkOMCL22,Xu2023H2ONetHN,
Wang2023HandGCATO3}, but metric scale remains implicit in their geometry and
translation. NVF \citep{Huang_2023_CVPR} instead supervises dense voting in a
camera frustum. HandDGP \citep{valassakis2024handdgp} makes global positioning
differentiable but reports a tension with root-relative accuracy, whereas HandOS
\citep{chen2025handos} jointly decodes geometry and translation without exposing
an independently supervised metric gauge. WiLoR improves hand localization, and
other systems exploit forearm or interaction context
\citep{Potamias2024WiLoRE3,millerdurai2026egoforce,
hasson20_handobjectconsist,spurr2020}. These methods advance global positioning,
but they do not provide an explicit instance-scale representation through which
metric supervision can coordinate with relative hand geometry. Consequently,
an aligned-pose gain need not reduce absolute error: plausible projections admit
different hand-size--depth pairs, and implicit metric variables can compensate
for one another during training.

\noindent\textbf{Root-relative HPE.}
Parametric, heatmap, and model-free methods respectively regress MANO parameters,
discretize 2.5D/3D coordinates, or directly regress joints and vertices
\citep{DBLP:journals/tog/RomeroTB17,DBLP:conf/cvpr/BaekKK19,hasson19_obman,
DBLP:conf/cvpr/BoukhaymaBT19,iqbal2018handpose,moon2020i2l,moon2020interhand,
DBLP:conf/cvpr/ChenLDZMXZG22,DBLP:conf/cvpr/GeRLXWCY19,DBLP:conf/cvpr/KulonGKBZ20}.
Transformers and detector-based systems introduce global context and structured
queries \citep{DBLP:conf/iccv/LinWL21,DBLP:conf/cvpr/LinWL21,
DBLP:conf/cvpr/PavlakosSRKFM24,DBLP:conf/nips/DongCG0T24,
DBLP:conf/cvpr/ShiWLRT22,Yang2023ExplicitBD,DBLP:conf/cvpr/SunWZYWWML00C24},
while specialized designs address efficiency, deformation, and occlusion
\citep{Yoshiyasu_2023_CVPR,Zhou_2024_CVPR,Saleem2024MaskHandGM,
Lu2024SPMHandSP,ren2025learning}. These methods achieve strong aligned accuracy,
but alignment removes translation and common scale; it therefore does not test
whether local geometry remains accurate when composed into metric camera space.

\begin{table*}[t]
\centering
{\small
\setlength{\tabcolsep}{5pt}
\begin{tabular}{@{}lccc@{}}
\toprule
Method & Metric-scale source & Scale feedback to pose & Pose representation \\ \midrule
Iqbal et al. & supplied/mean bone length & post-hoc & 2.5D joint heatmaps \\
CMR & implicit in root-relative geometry & post-hoc & semantic feature aggregation \\
NVF & direct/test reference & direct camera-space & dense frustum voting field \\
HandDGP & implicit in root-relative geometry & differentiable (trade-off) & image-to-geometry regression \\
HandOS & implicit in geometry and translation & one-way pose-to-translation & 2D-joint and 3D-vertex queries \\
EgoForce & hand--arm shape/context & joint camera-space & hand--arm parameter features \\
\textbf{ScaleHP} & \textbf{predicted instance scalar} & \textbf{bidirectional scale--pose} & \textbf{sparse joint and scale tokens} \\ \bottomrule
\end{tabular}}
\caption{Metric-scale handling in representative camera-space HPE.
``Scale feedback to pose'' summarizes how metric/global objectives communicate
with pose features. The final column lists the intermediate pose representation, in which NVF's voting field is dense.}
\label{tab:method_positioning}
\end{table*}

\noindent\textbf{Metric scale learning.}
Dense metric prediction couples a single global gauge to every local depth or
point-map element. ScaleDepth \citep{song2024scaledepth} explicitly separates
scene-scale queries from normalized relative-depth bin queries so that the dense
relative branch can model structure without absorbing cross-scene depth-range
variation. MoGe-2 \citep{wang2025moge} reaches the same conclusion more directly:
its entangled metric point-map baseline is suboptimal, while decoupling a global
CLS-derived scale from an affine-invariant point map better preserves relative
geometry. Other metric-depth methods use adaptive bins, canonical-camera rays,
or scale-aware dense representations \citep{bhat2021adabins,li2024binsformer,
yin2023metric3d,guizilini2023towards,zhang2025metrichmr}. The shared lesson is
not merely that scale can be predicted, but that folding a global scale into a
high-dimensional local representation can make metric and relative objectives
compete: the global quantity must determine metric extent, yet entangling it with
dense local elements can distort their relative geometry. Hand pose provides a
more compact interface because its geometry admits a sparse, semantically indexed
joint representation. ScaleHP exploits this property by keeping scale and
unit-scale joint tokens distinct while allowing them to interact, so global
metric reasoning and local geometric learning can be optimized jointly.

\section{Method}

\subsection{Why Scale Matters}
\label{sec:why_scale}

Table~\ref{tab:method_positioning} contrasts how camera-space methods represent
and expose metric scale. We use \emph{scale feedback to pose} for the direction
in which metric/global supervision affects pose features. Prior choices leave
scale supplied externally or distributed across high-dimensional geometry and
translation.

Let the ground-truth camera-space pose be
$\mathbf Q=s(\mathbf J+\mathbf t)$, where $\mathbf J$ is unit-scale
root-relative geometry and $\mathbf t$ is normalized root translation. With
$\hat s=s+\Delta s$, $\hat{\mathbf J}=\mathbf J+\Delta\mathbf J$, and
$\hat{\mathbf t}=\mathbf t+\Delta\mathbf t$, the exact error is
\begin{equation}
\begin{split}
\hat{\mathbf Q}-\mathbf Q={}&
\underbrace{s\Delta\mathbf J}_{\text{(a) relative geometry}}+
\underbrace{s\Delta\mathbf t}_{\text{(b) root localization}}\\
&+\underbrace{\Delta s(\mathbf J+\mathbf t)
+\Delta s(\Delta\mathbf J+\Delta\mathbf t)}_{\text{(c) scale-mediated coupling}}.
\end{split}
\label{eq:error_decomposition}
\end{equation}
Since $\lVert\mathbf t\rVert\gg\lVert\mathbf J\rVert$ in camera space,
$\lVert\Delta s(\mathbf J+\mathbf t)\rVert\approx
|\Delta s|\lVert\mathbf t\rVert$: even a small scale error can induce a large
depth displacement. Moreover, the last term couples global scale to both local
geometry and root positioning. This motivates exposing $s$ as a separately
supervised instance variable rather than leaving it implicit in a dense pose or
translation representation.

\noindent\textbf{Explicit scale prediction.}
We factor each metric joint as
$\mathbf Q^{cam}=s(\mathbf J^{rel}+\mathbf t^{norm})$, where
$\mathbf J^{rel}$ is root-relative and unit-scale. Perspective correspondences
recover $\mathbf t^{norm}$ under known intrinsics, but cannot determine $s$:
scaling both hand and translation leaves the projection unchanged. Hence $s$
is learned from metric supervision and remains a data-driven monocular
prior.

Let $\mathbf J^{M*}$ denote the $J$ ground-truth joint annotations in metric
camera coordinates and let $r$ index the root. The
network does not regress $\mathbf J^{M*}$ directly. Instead, it receives the
two targets induced by these annotations: projected pixel joints
$\mathbf J^{2D*}$ and unit-scale root-relative joints $\mathbf J^{3D*}$.
Motivated by their stable long-axis
anatomy \citep{Varu2015DeterminationOS,Rastogi2008EstimationOS,
Manning2002DigitRA},
let $\{l_i\}_{i=0}^{N-1}$ be the metric bone lengths in $\mathbf J^{M*}$ along
the index, middle, and ring fingers (IMR). We define
\begin{equation}
\begin{aligned}
s^*&=\frac{1}{N}\sum_{i=0}^{N-1}l_i,\\
\mathbf J^{2D*}&=\pi(\mathbf J^{M*},\mathbf K_{cam}),\\
\mathbf J^{3D*}&=\frac{\mathbf J^{M*}-\mathbf J^{M*}_{r}}{s^*}.
\end{aligned}
\label{eq:scale_target}
\end{equation}
Thus the architecture outputs $\mathbf J^{2D}\!\rightarrow\!\mathbf J^{2D*}$,
$\mathbf J^{3D}\!\rightarrow\!\mathbf J^{3D*}$, and
$s\!\rightarrow\!s^*$; the analytical transfer recomposes them into metric
camera space. While this target focuses on IMR fingers, it does not rule out scale
evidence from other fingers; the ablation study and supplementary material provide
further details. A scale token aggregates joint layout, unit-scale geometry, and
image appearance, allowing metric supervision to affect pose features before
calibrated positioning.

\subsection{Architecture}
\label{sec:arch}
\noindent\textbf{Overview.} Figure~\ref{fig:overview} summarizes ScaleHP, which
comprises a frozen DETR-style detector, a
jointly trained metric-aware 2D--3D decoder, and a parameter-free analytical
module. A \textbf{scale token} is updated jointly with hand-related 2D semantic
and 3D geometric queries and cross-attends to image features. The decoder predicts
an instance metric gauge alongside unit-scale pose, after which the analytical
module solves normalized translation under known intrinsics. The complete system
performs one-pass inference without iterative registration.

\subsubsection{DETR-Like Detector.}

Frozen Grounding DINO \citep{Liu2023GroundingDM} with prompt \textit{hand}
fuses visual/text features and iteratively updates content queries and 2D
references through deformable decoding \citep{DBLP:conf/iclr/ZhuSLLWD21}:
\begin{equation}
\mathbf F_e=\mathcal E(\mathcal B_v(\mathbf I),\mathcal B_t(\text{``hand''})).
\label{eq:detector_encoder}
\end{equation}
\begin{equation}
(\mathbf F_{GD},\mathbf Q,\mathbf R,\mathbf B,\mathbf C)
=\mathcal D_{det}(\mathbf F_e).
\label{eq:detector_interface}
\end{equation}
Here $\mathbf F_{GD}$ contains multi-scale features and
$(\mathbf Q,\mathbf R,\mathbf B,\mathbf C)$ are instance queries, references,
boxes, and confidences. After score filtering and NMS, the retained set is
$\{(\tilde{\mathbf Q}_k,\tilde{\mathbf R}_k,
\tilde{\mathbf B}_k)\}_{k=1}^{K}$. The hand decoder processes all $K$ instances;
the complete interface is in the supplementary material.

\subsubsection{Metric 2D-3D Decoder.}

Each retained instance is expanded into $J=21$ semantically indexed 2D and 3D
joint queries using learnable embeddings $\mathbf E_J$; a hand-level scale
token $\mathbf q^s$ is initialized from the same instance context. We suppress
the instance index $k$ below because the same decoder is applied independently
to every retained hand:
\begin{equation}
\mathbf Q^{2D}_0=\mathbf Q^{3D}_0=\tilde{\mathbf Q}+\mathbf E_J,
\qquad \mathbf q^s_0=\tilde{\mathbf Q}.
\label{eq:decoder_init}
\end{equation}
At layer $l$, the scale token and both joint streams exchange information through
global self-attention:
\begin{equation}
(\mathbf q_l^{s\prime},\mathbf Q_l^{2D},\mathbf Q_l^{3D})
=\operatorname{SA}_l(\mathbf q_{l-1}^{s},\mathbf Q_{l-1}^{2D},
\mathbf Q_{l-1}^{3D}).
\label{eq:decoder_self_attention}
\end{equation}
This interaction is reciprocal: the scale token aggregates all joint layouts and
normalized structures, and every joint attends to scale-supervised context.
Repeated layers therefore refine both directions before prediction.

Each stream then samples multi-scale appearance around its reference points:
\begin{equation}
\begin{aligned}
\mathbf q^s_l&=\operatorname{DefAttn}^s_l
(\mathbf q_l^{s\prime},\mathbf F_{GD},\mathbf R^s_{l-1}),\\
\mathbf Q^d_l&=\operatorname{DefAttn}^d_l
(\mathbf Q^d_l,\mathbf F_{GD},\mathbf R^d_{l-1}),\\
\mathbf R^d_l&=\mathbf R^d_{l-1}+\operatorname{FFN}^d_R(\mathbf Q^d_l),
\end{aligned}
\label{eq:decoder_update}
\end{equation}
for $d\in\{2D,3D\}$; the scale reference uses
$\mathbf R_l^s=\mathbf R_{l-1}^s+\operatorname{FFN}^s_R(\mathbf q_l^s)$.
Deformable attention adds local appearance and occupancy evidence, while
reference refinement follows the evolving estimates. Finally,
\begin{equation}
\mathbf J^{2D}=\operatorname{FFN}_{2D}(\mathbf Q^{2D}_L),
\qquad
\mathbf J^{3D}=\operatorname{FFN}_{3D}(\mathbf Q^{3D}_L).
\label{eq:pose_prediction_heads}
\end{equation}
\begin{equation}
s=\operatorname{FFN}_{s}(\mathbf q^s_L).
\label{eq:scale_prediction_head}
\end{equation}
predict pixel coordinates, unit-scale root-relative joints, and the metric
scalar supervised by Eq.~\ref{eq:scale_target}, respectively.

\subsubsection{Transfer to Real Metric.}
\label{sec:transfer}
The training-free analytical solver receives $\mathbf J^{2D}$,
$\mathbf J^{3D}$, $s$, and known intrinsics. Let $p$ be the root joint and let
$\mathbf A$ stack
$[u_j-u_p,\,v_j-v_p]^\top$ for $j\ne p$. Perspective projection gives the
corresponding right-hand side
\begin{equation}
\mathbf b=\left[
\begin{aligned}
f_xx_j-(u_j-c_x)z_j,\\
f_yy_j-(v_j-c_y)z_j
\end{aligned}
\right]_{j\ne p}^{\top}.
\label{eq:solver_rhs}
\end{equation}
The normalized translation is recovered without scale as
\begin{equation}
\begin{gathered}
t_z^{norm}=\mathbf A^{\dagger}\mathbf b,\\
t_x^{norm}=\frac{u_p-c_x}{f_x}t_z^{norm},\qquad
t_y^{norm}=\frac{v_p-c_y}{f_y}t_z^{norm}.
\end{gathered}
\label{eq:compact_solver}
\end{equation}
Then $\mathbf Q^{cam}=s(\mathbf J^{3D}+\mathbf t^{norm})$ composes the metric
prediction without learned translation regression or iterative registration.
The supplementary material gives the full projection derivation.

\subsection{Loss Functions}
\label{sec:loss_scope}

ScaleHP is jointly optimized with 2D, unit-scale 3D, scale, and projection
supervision. For ground truth denoted by $(*)$, the direct terms are
\begin{equation}
\begin{aligned}
\mathcal L^{2D}&=\lVert\mathbf J^{2D}-\mathbf J^{2D*}\rVert_1,\\
\mathcal L^{3D}&=\lVert\mathbf J^{3D}-\mathbf J^{3D*}\rVert_1,\\
\mathcal L^s&=(s-s^*)^2.
\end{aligned}
\label{eq:direct_losses}
\end{equation}
with an additional OKS loss on 2D joints. Following \citet{chen2025handos}, the
differentiable solve $\hat{\mathbf t}^{norm}=\operatorname{LS}
(\mathbf J^{2D},\mathbf J^{3D},\mathbf K_{cam})$ supplies weak reprojection
supervision:
\begin{equation}
\begin{aligned}
\hat{\mathbf{J}}^{proj}&=\pi(\mathbf{J}^{3D}+\hat{\mathbf{t}}^{norm},
\mathbf{K}_{cam}),\\
\mathcal{L}^{proj}_{2D}&=\left\Vert \hat{\mathbf{J}}^{proj} -
\mathbf{J}^{2D*}\right\Vert_1,\\
\mathcal{L}^{proj}_{OKS}&=1-\mathit{OKS}(\hat{\mathbf{J}}^{proj},
\mathbf{J}^{2D*}).
\end{aligned}
\label{eq:projection_losses}
\end{equation}
The solver receives no separate ground-truth camera-space coordinate loss;
joint learning refers to scale/pose representations and reprojection, not full
camera-space supervision through every component.

\section{Experiments}

\noindent\textbf{Setup.}
We evaluate FreiHand \citep{zimmermann2019freihand}, DexYCB
\citep{chao2021dexycb}, and HO3Dv3 \citep{Hampali2019HOnnotateAM}; HInt,
COCO-WholeBody, CompHand, H2O3D and OneHand10K appear only in the combined-data protocol.
Frozen Grounding DINO 1.5 processes full images at a 1280-pixel long edge. We
train 45 epochs with Adam, batch size 16, $10^{-4}$ initial learning rate, and
cosine decay on eight A100 GPUs. CS-MPJPE is unaligned camera-space error,
R-MPJPE root-aligns the prediction to the ground-truth root, and PA-MPJPE applies
a similarity Procrustes alignment, following established camera-space and
aligned-pose protocols \citep{Huang_2023_CVPR,valassakis2024handdgp,
millerdurai2026egoforce}. Training details are supplementary.

\subsection{Main Results}

CS-MPJPE is primary because it applies neither alignment nor post-hoc scale
correction. We separate the evaluation into an in-dataset FreiHand comparison
and cross-dataset transfer. Table~\ref{tab:freihand_main} reports results from
the original FreiHand protocols;  Table~\ref{tab:cross_dataset} instead
applies the same FreiHand-trained checkpoint to DexYCB and HO3Dv3 because baseline
methods do not provide compatible checkpoints trained on those target datasets.

\begin{table}[t]
\centering
{\small
\setlength{\tabcolsep}{5pt}
\begin{tabular}{@{}lcc@{}}
\toprule
Method & CS-MPJPE $\downarrow$ & PA-MPJPE $\downarrow$ \\
\midrule
CMR-PG & 48.8 & 6.9 \\
HandDGP & 46.3 & 7.4 \\
MobRecon & 50.2 & 5.7 \\
NVF & 42.4 & - \\
METRO & - & 6.7 \\
MaskHand & - & 5.5 \\
Hamba & - & 5.7 \\
HandOS & - & \textbf{5.0} \\
\textbf{ScaleHP} & \textbf{35.8} & \textbf{5.0} \\ \bottomrule
\end{tabular}}
\caption{FreiHand comparison under the protocols reported by the respective
methods (mm). CS-MPJPE measures unaligned camera-space accuracy; PA-MPJPE applies
Procrustes alignment. NVF's reported 42.4\,mm uses the benchmark reference-bone
scale (47.2\,mm without it). ScaleHP uses RGB and known intrinsics without
test-time ground-truth scale.}
\label{tab:freihand_main}
\end{table}

\begin{table}[t]
\centering
{\small
\setlength{\tabcolsep}{5pt}
\begin{tabular}{@{}lcc@{}}
\toprule
Method & DexYCB $\downarrow$ & HO3Dv3 $\downarrow$ \\
\midrule
CMR + oracle global scale & 183.2 & 152.3 \\
HandDGP + oracle global scale & 222.1 & 132.6 \\
\textbf{ScaleHP} & \textbf{136.3} & \textbf{50.7} \\ \bottomrule
\end{tabular}}
\caption{Cross-dataset CS-MPJPE using FreiHand-trained checkpoints (mm).
The oracle global scale is a test-set-optimal dataset-level rescaling introduced
only to give CMR and HandDGP, which do not expose an instance scale, a favorable
transfer calibration. ScaleHP uses neither target-dataset training nor
test-time ground-truth scale.}
\label{tab:cross_dataset}
\end{table}

\begin{figure}[!t]
\centering
\setlength{\tabcolsep}{1pt}
\begin{tabular}{>{\centering\arraybackslash}m{0.22\columnwidth}
                >{\centering\arraybackslash}m{0.34\columnwidth}
                >{\centering\arraybackslash}m{0.34\columnwidth}}
\textbf{Input} & \textbf{Metric Space} & \textbf{Top view} \\[2pt]
\includegraphics[width=\linewidth]{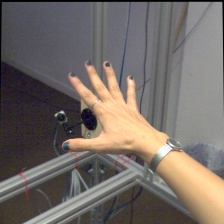} &
\includegraphics[width=\linewidth]{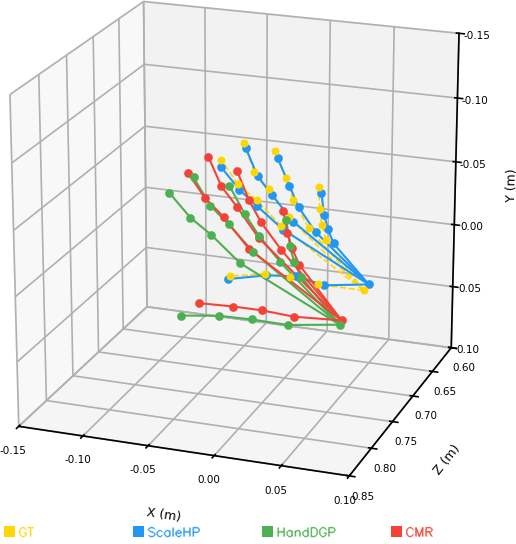} &
\includegraphics[width=\linewidth]{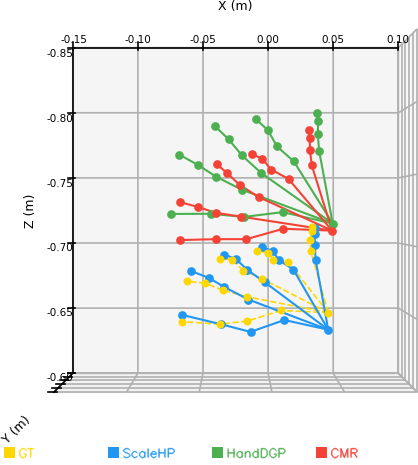} \\[3pt]
\includegraphics[width=\linewidth]{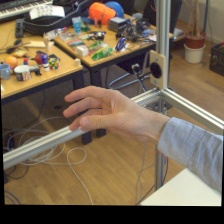} &
\includegraphics[width=\linewidth]{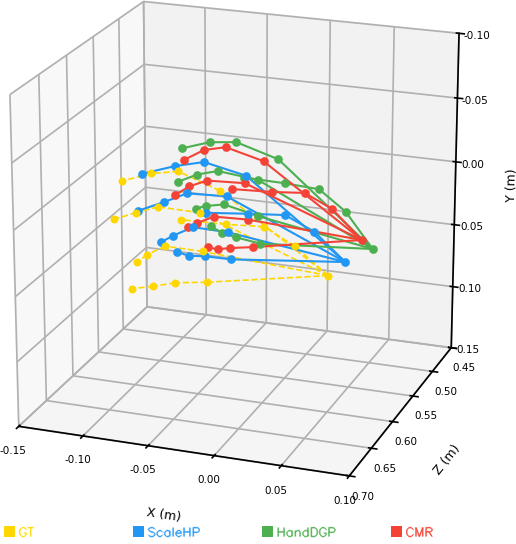} &
\includegraphics[width=\linewidth]{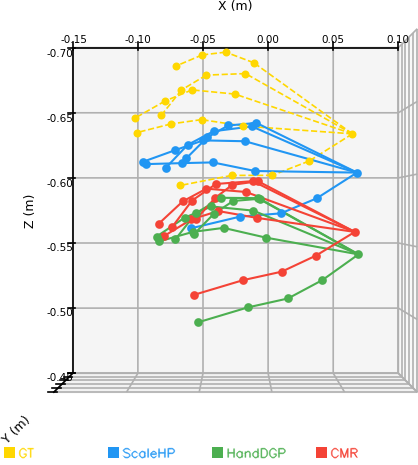} \\[3pt]
\includegraphics[width=\linewidth]{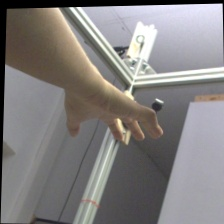} &
\includegraphics[width=\linewidth]{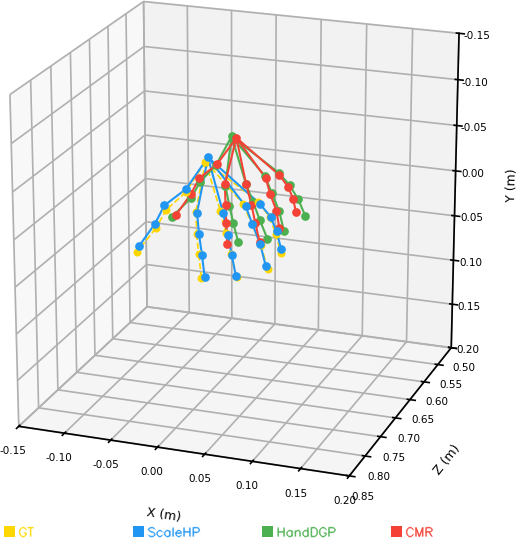} &
\includegraphics[width=\linewidth]{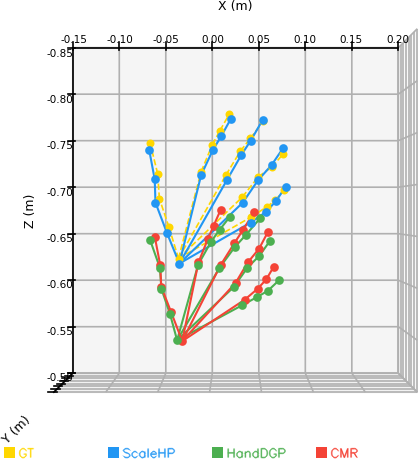} \\[3pt]
\includegraphics[width=\linewidth]{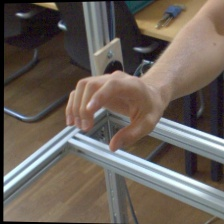} &
\includegraphics[width=\linewidth]{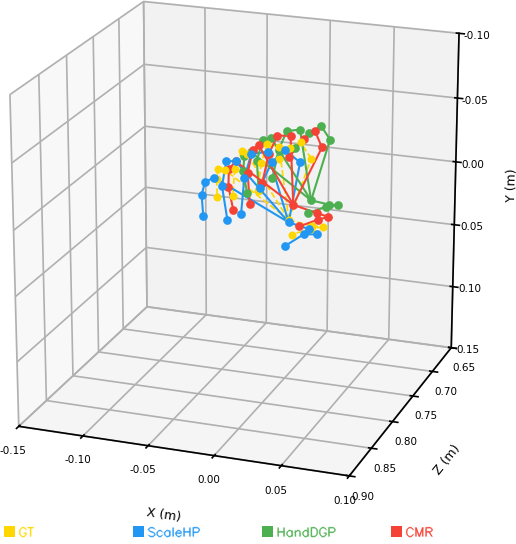} &
\includegraphics[width=\linewidth]{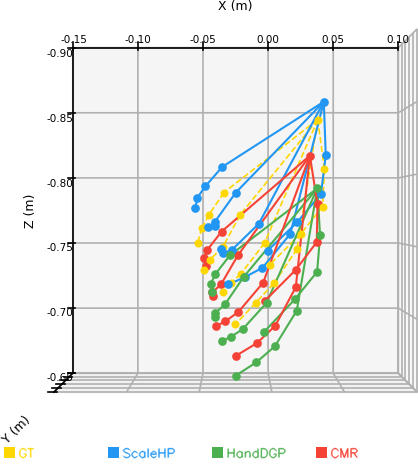}
\end{tabular}
\caption{Main FreiHand camera-space comparisons on four samples. Yellow: GT; blue:
ScaleHP; green: HandDGP; red: CMR. The metric space and top views show that ScaleHP remains closest to GT
in absolute translation.}
\label{fig:qualitative_main}
\end{figure}

On FreiHand, ScaleHP reduces CS-MPJPE from the best listed 42.4\,mm to
35.8\,mm (15.6\%) while tying the best PA-MPJPE of 5.0\,mm. DexYCB and HO3Dv3
in Table~\ref{tab:cross_dataset} are transfer tests, not target-dataset
retraining. CMR and HandDGP lack standalone scale outputs, so their rows receive
an external test-set-optimal dataset scalar; this oracle is needed to avoid
penalizing them for a dataset-level gauge mismatch when transferring from
FreiHand. Even with that favorable calibration, ScaleHP reduces CS-MPJPE over
the strongest available baseline by 25.6\% on DexYCB and 61.8\% on HO3Dv3.
The within-dataset analyses below separate the three error sources.

\subsubsection{Qualitative Results.}
Figure~\ref{fig:qualitative_main} compares four FreiHand samples with metric GT.
The metric space and top views respectively reveal absolute placement
and depth offsets. ScaleHP follows GT more closely than HandDGP and CMR,
particularly along global depth; additional views are in the supplementary material.

Figure~\ref{fig:qualitative_wild} shows four in-the-wild groups under occlusion,
object interaction, and viewpoint changes. Although these images lie outside the
training benchmarks, ScaleHP produces consistent metric reconstructions, indicating
strong generalization to unconstrained, out-of-dataset imagery.

\begin{figure*}[t]
\centering
\setlength{\tabcolsep}{0pt}
\begin{tabular}{@{}*{8}{>{\centering\arraybackslash}m{0.125\linewidth}}@{}}
\shortstack{\textbf{Original}\\\textbf{Image}} &
\shortstack{\textbf{Box}\\\textbf{and Pose}} &
\shortstack{\textbf{Metric}\\\textbf{Space}} &
\shortstack{\textbf{Top}\\\textbf{View}} &
\shortstack{\textbf{Original}\\\textbf{Image}} &
\shortstack{\textbf{Box}\\\textbf{and Pose}} &
\shortstack{\textbf{Metric}\\\textbf{Space}} &
\shortstack{\textbf{Top}\\\textbf{View}} \\[-1pt]
\includegraphics[width=\linewidth]{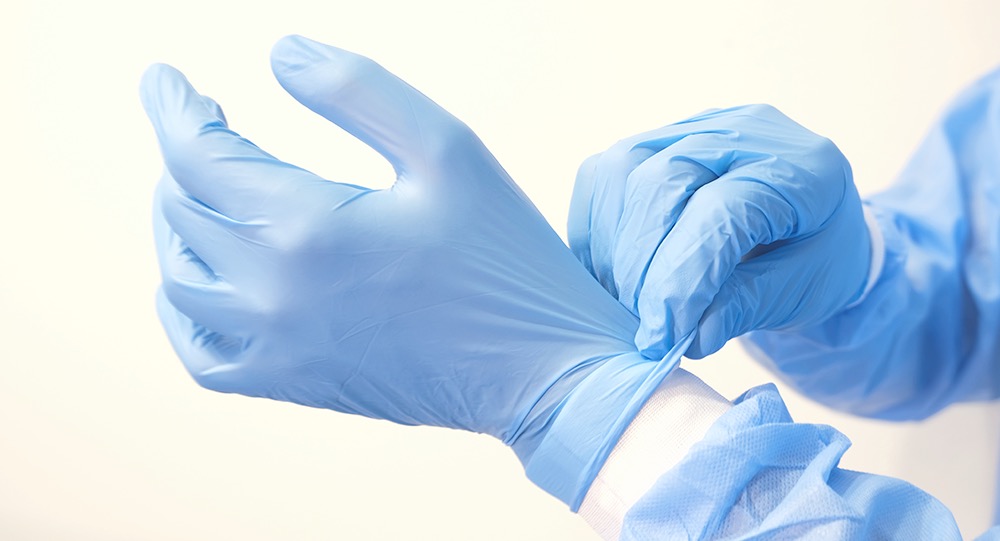} &
\includegraphics[width=\linewidth]{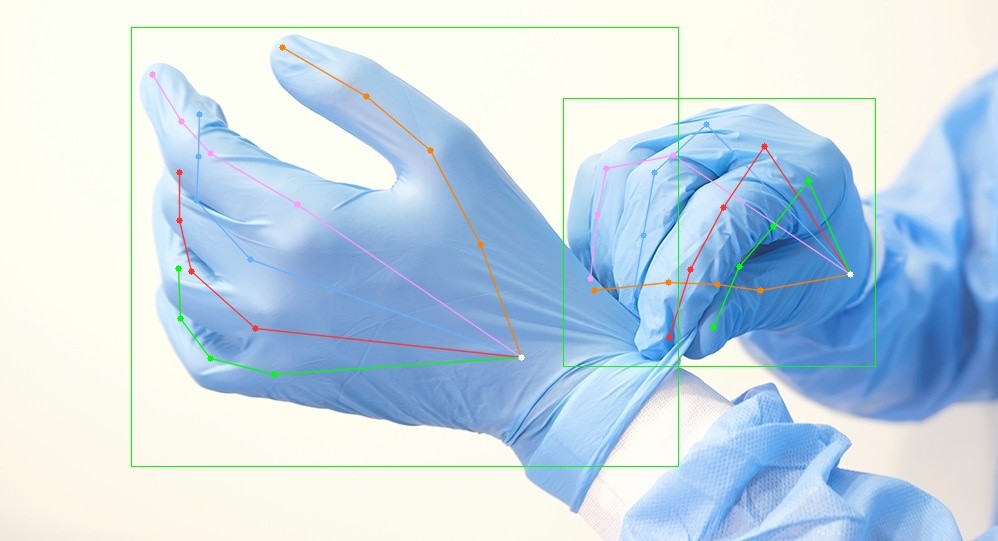} &
\includegraphics[width=\linewidth]{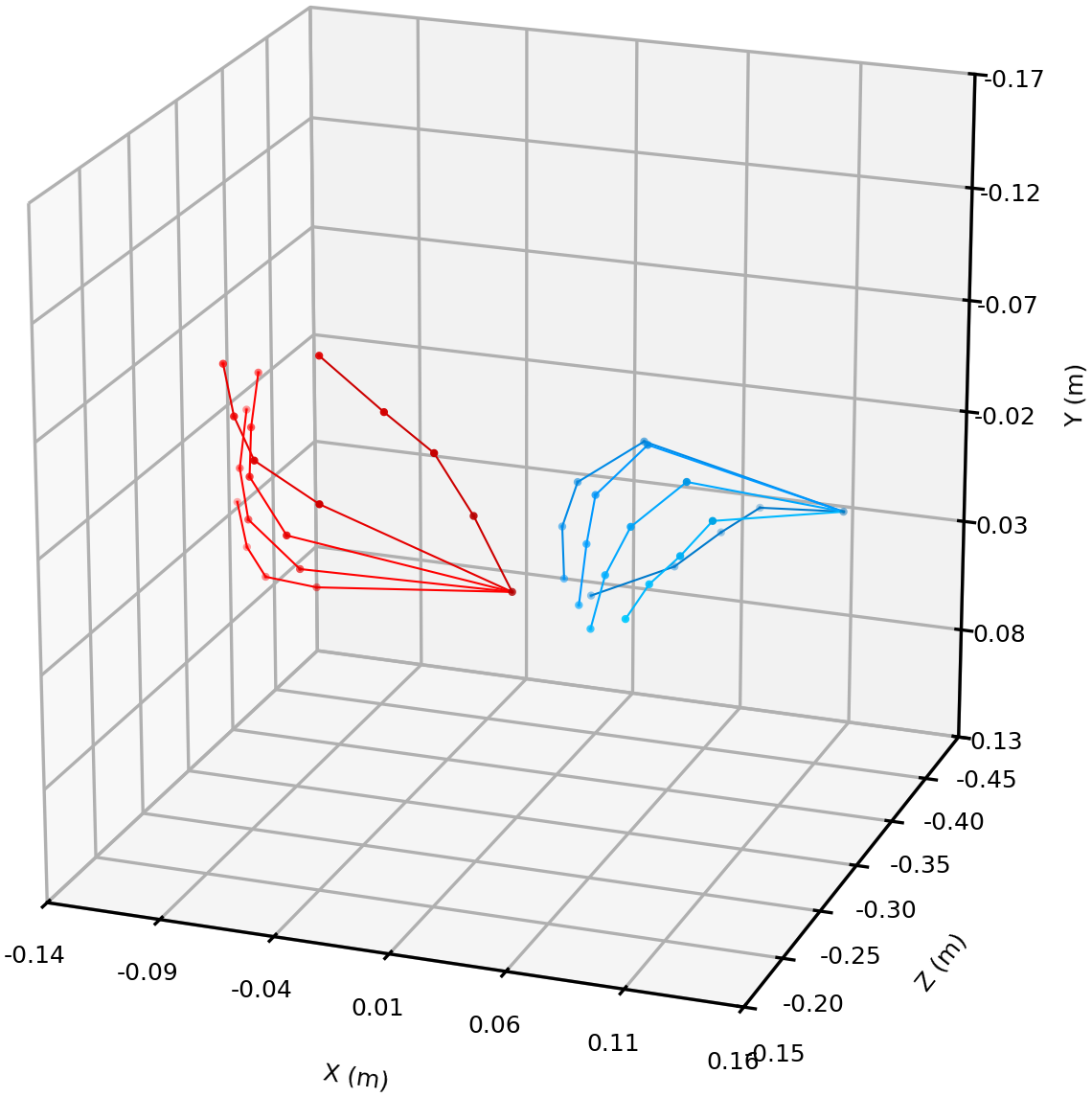} &
\includegraphics[width=\linewidth]{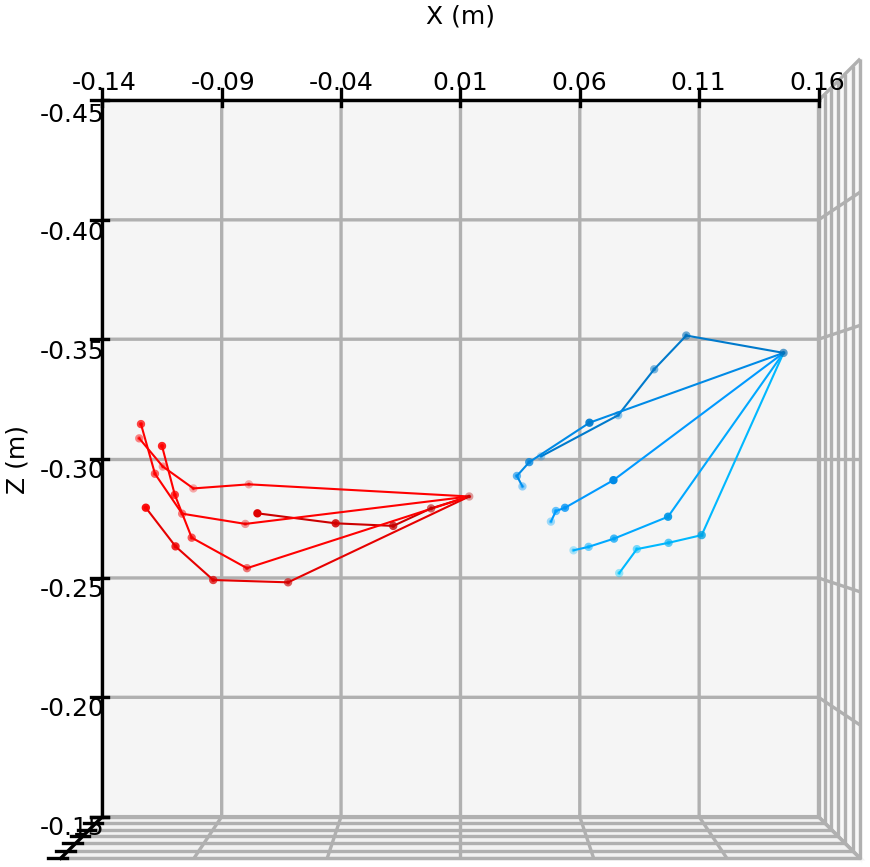} &
\includegraphics[width=\linewidth]{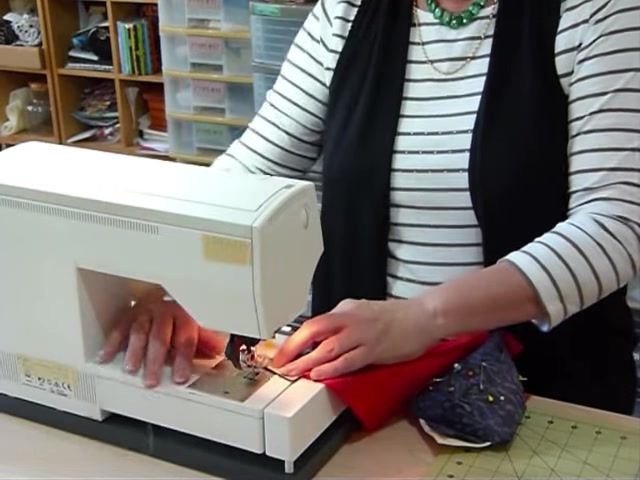} &
\includegraphics[width=\linewidth]{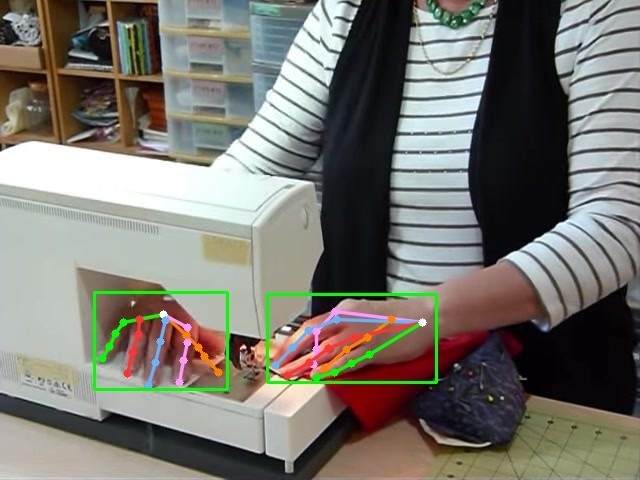} &
\includegraphics[width=\linewidth]{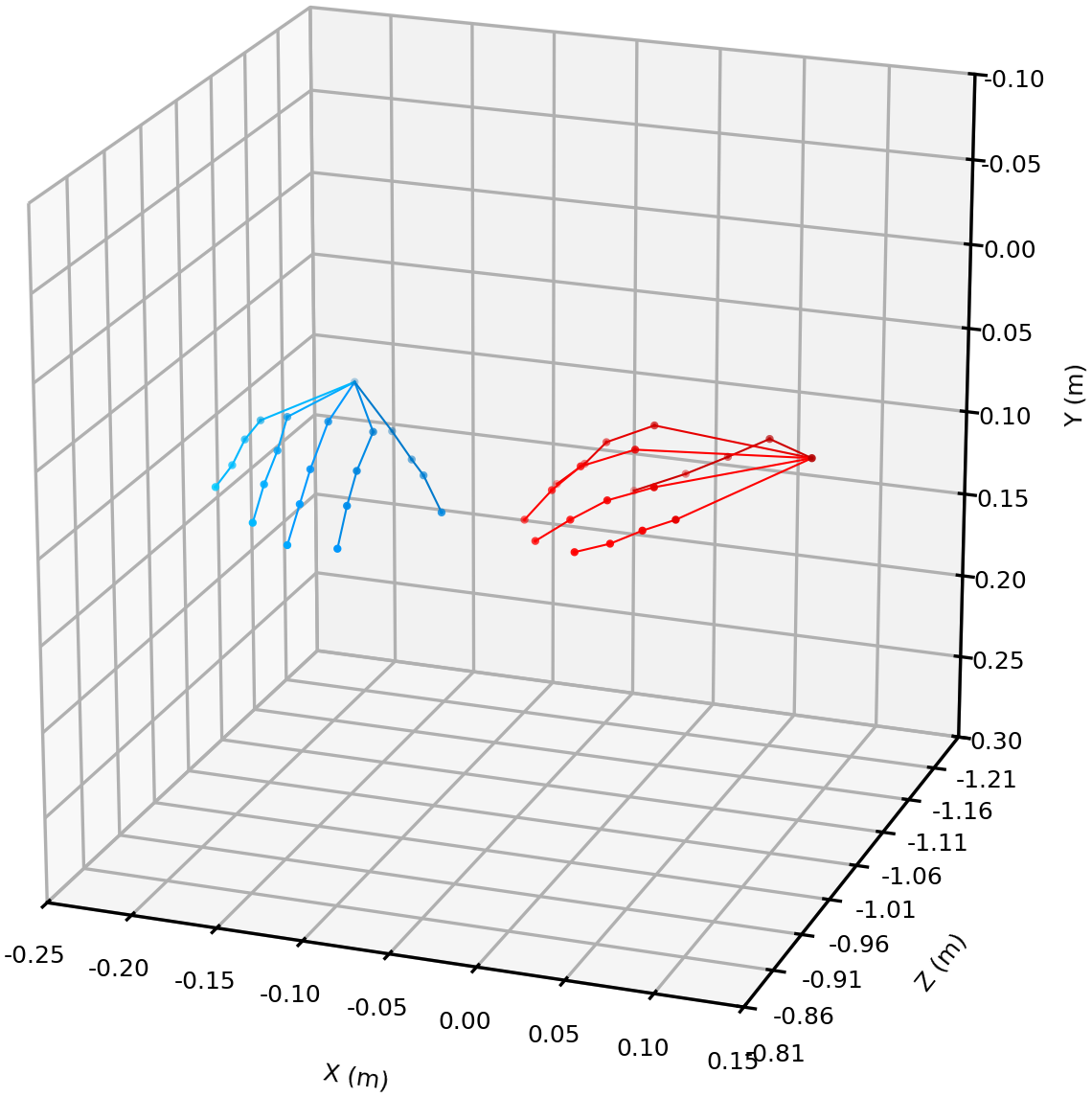} &
\includegraphics[width=\linewidth]{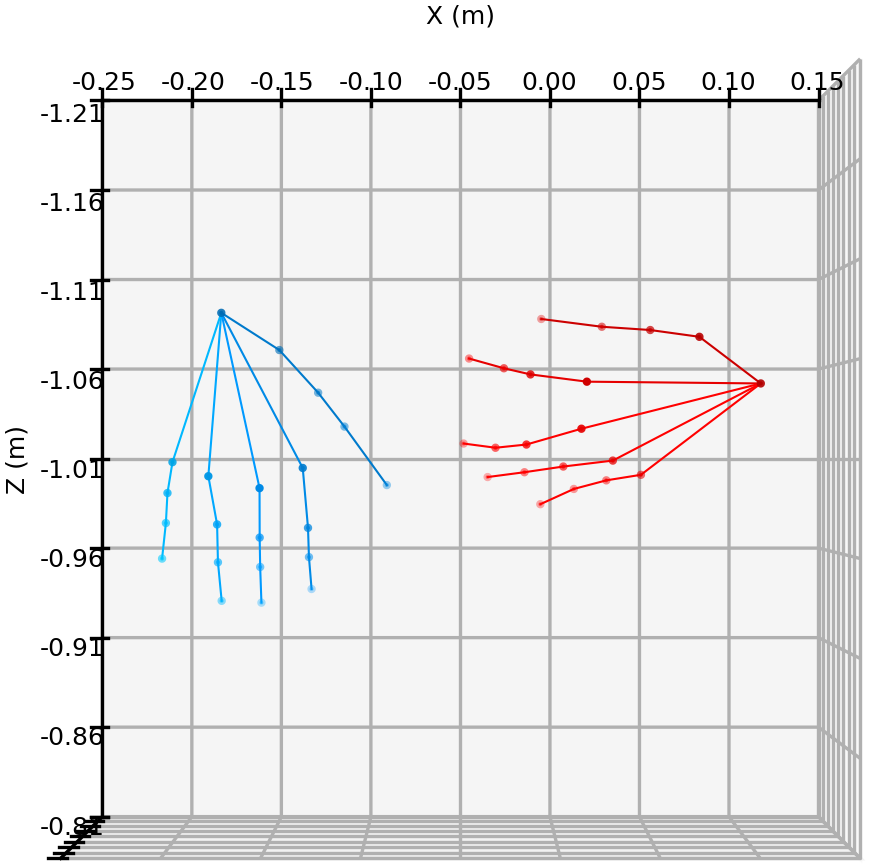} \\[3pt]
\includegraphics[width=\linewidth]{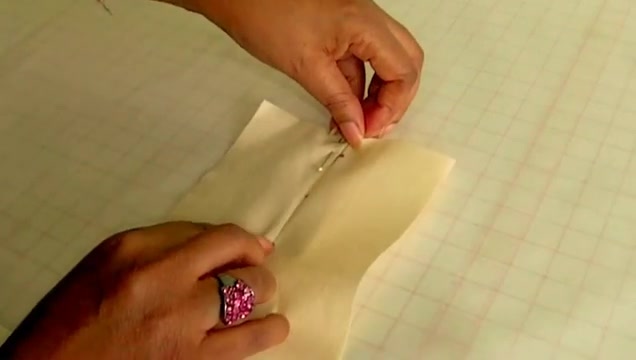} &
\includegraphics[width=\linewidth]{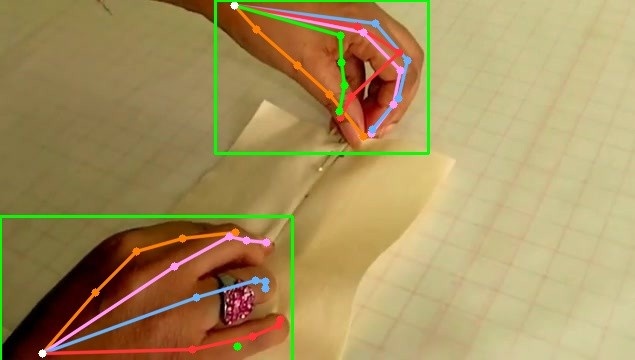} &
\includegraphics[width=\linewidth]{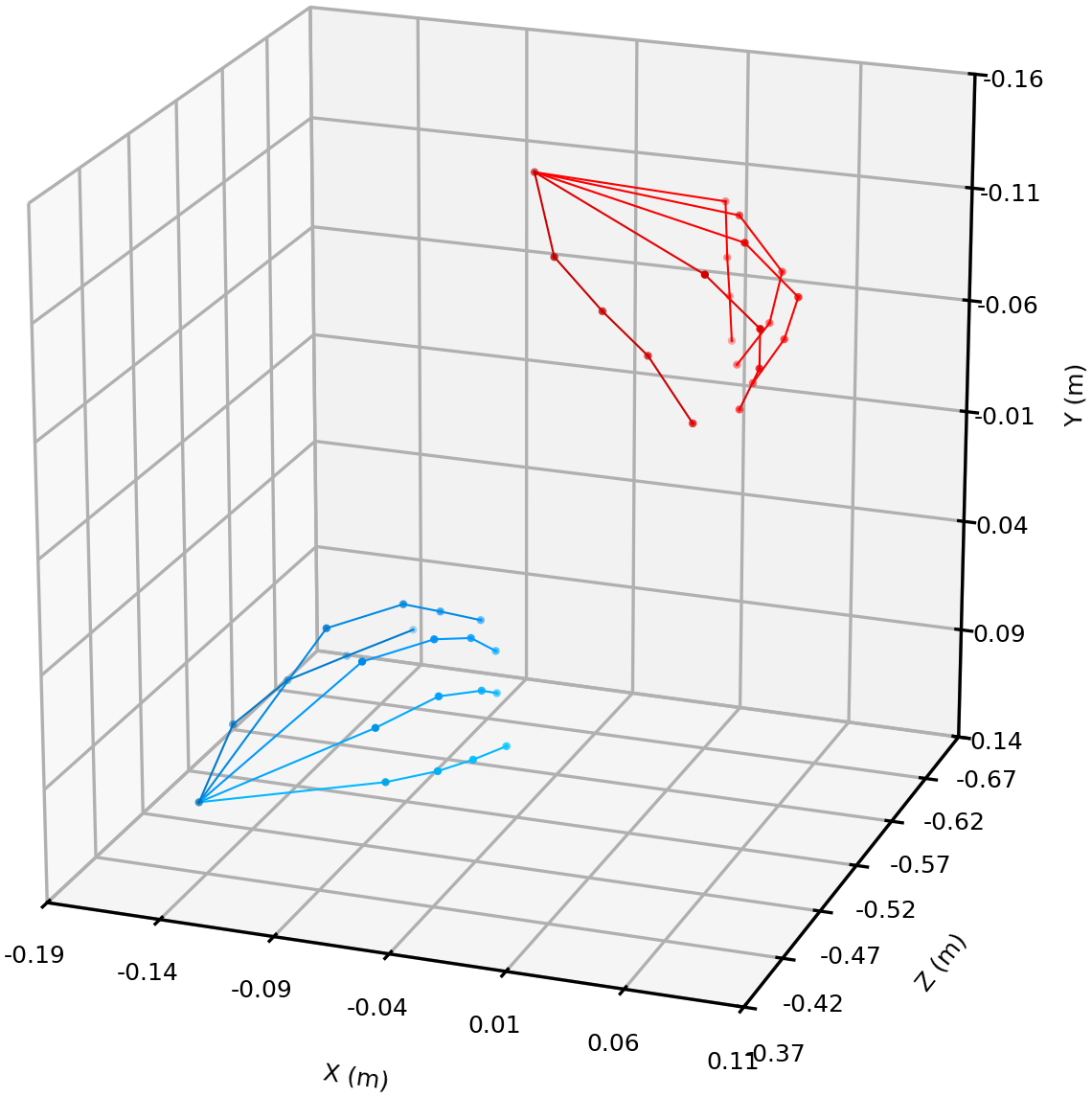} &
\includegraphics[width=\linewidth]{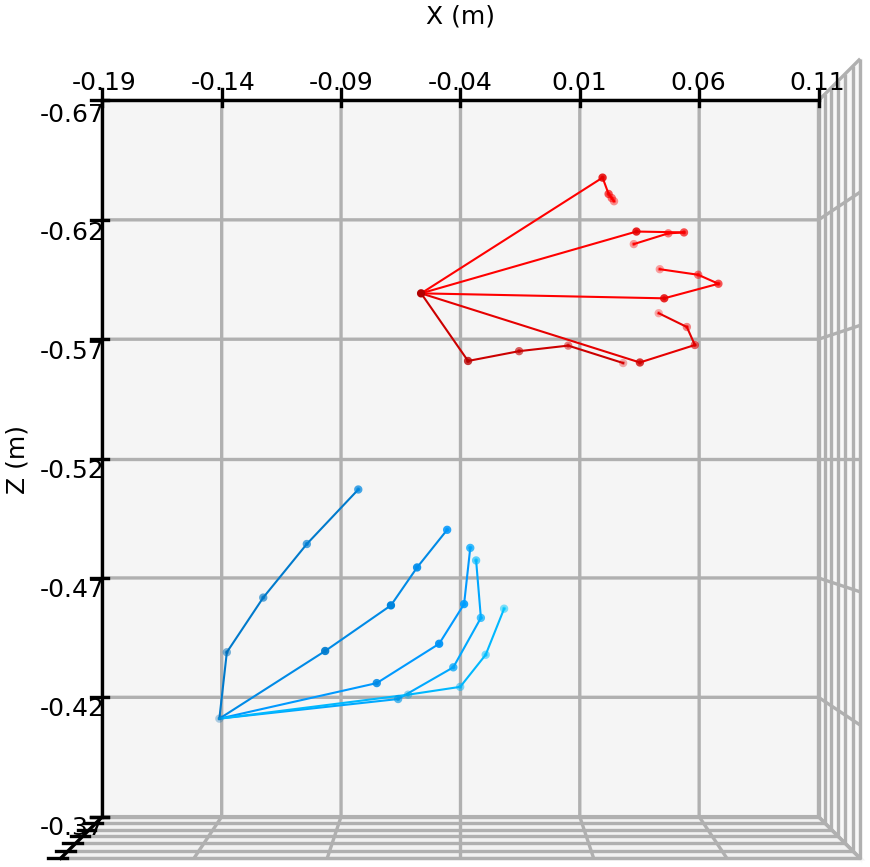} &
\includegraphics[width=\linewidth]{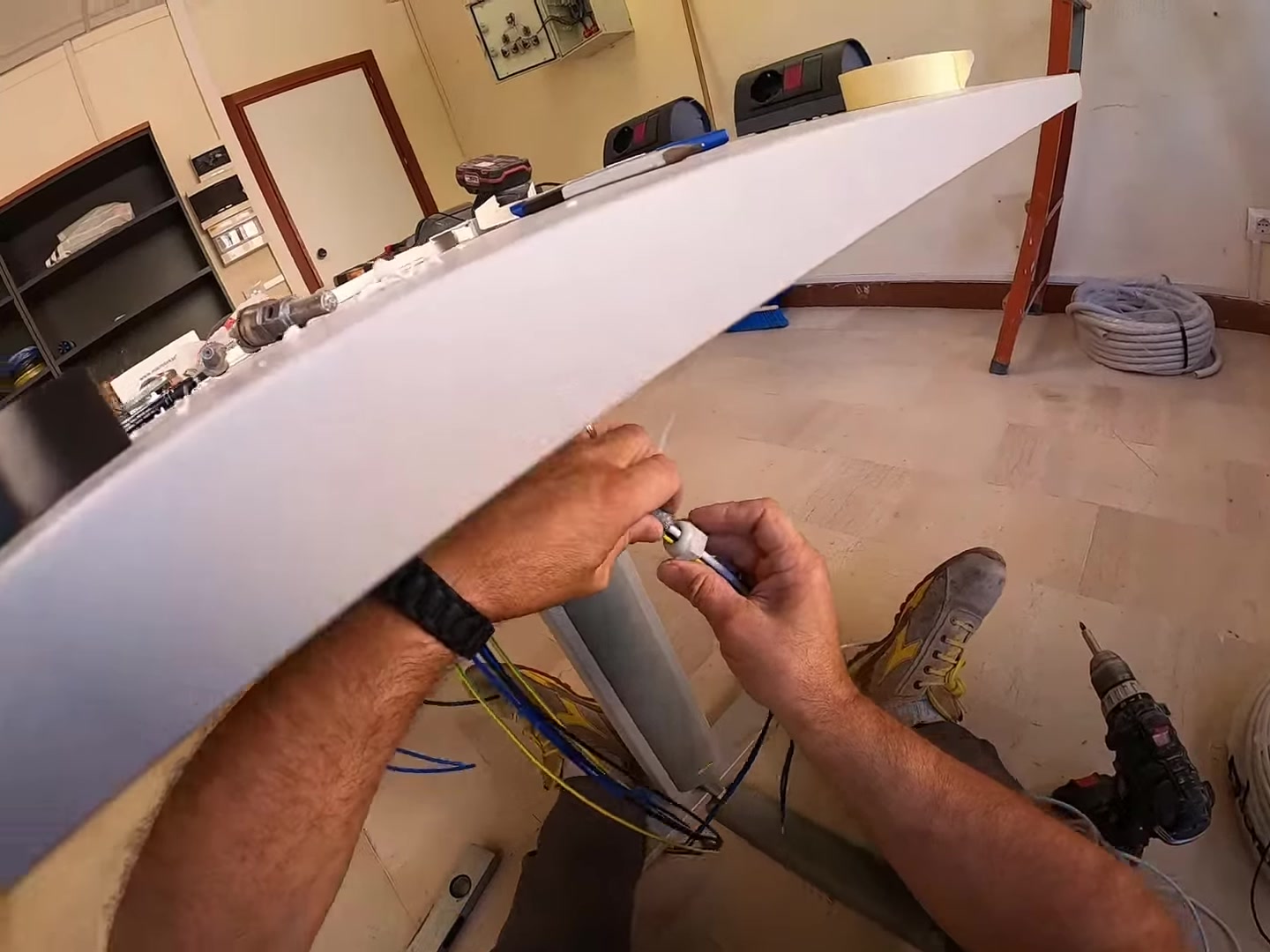} &
\includegraphics[width=\linewidth]{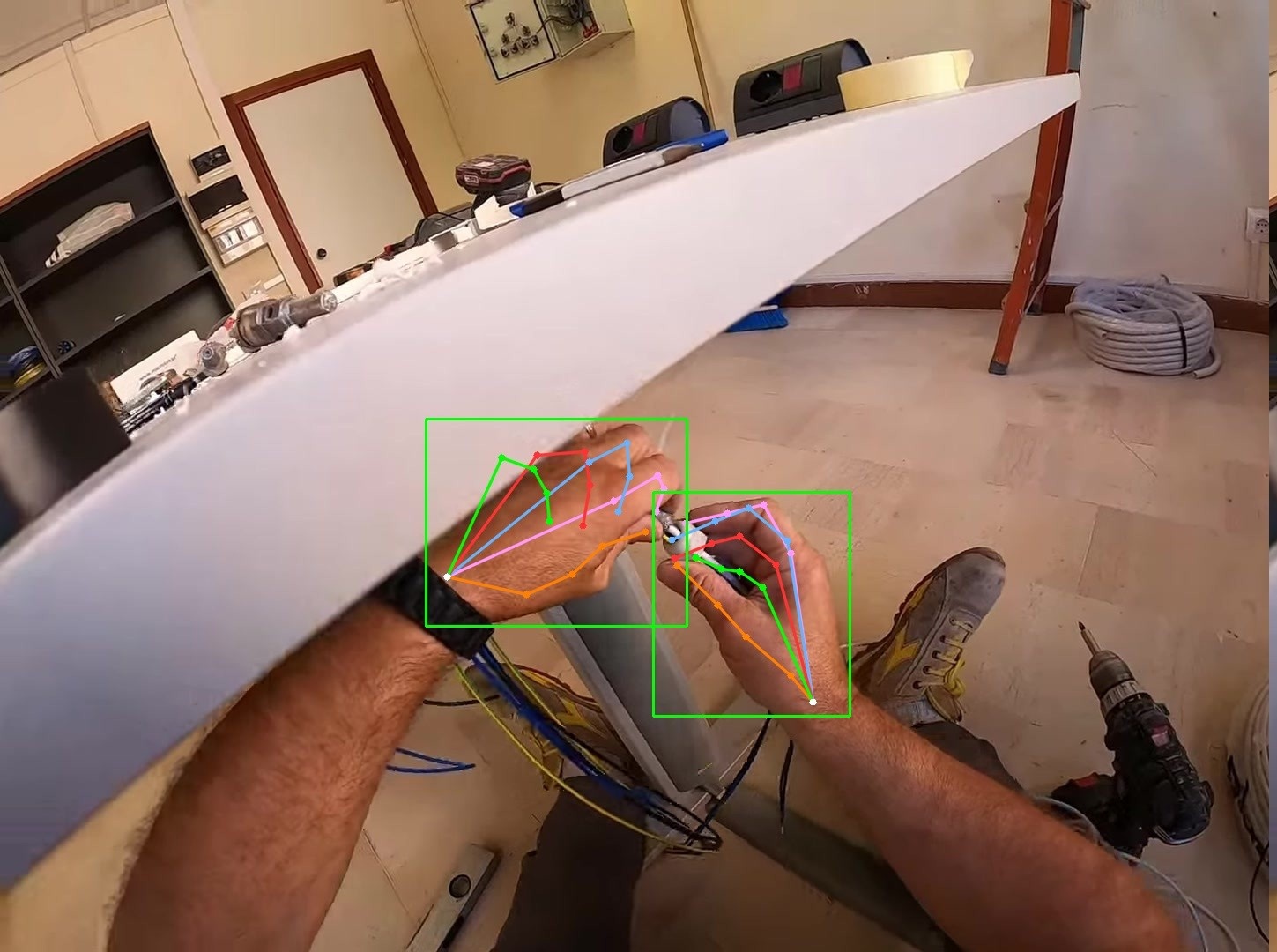} &
\includegraphics[width=\linewidth]{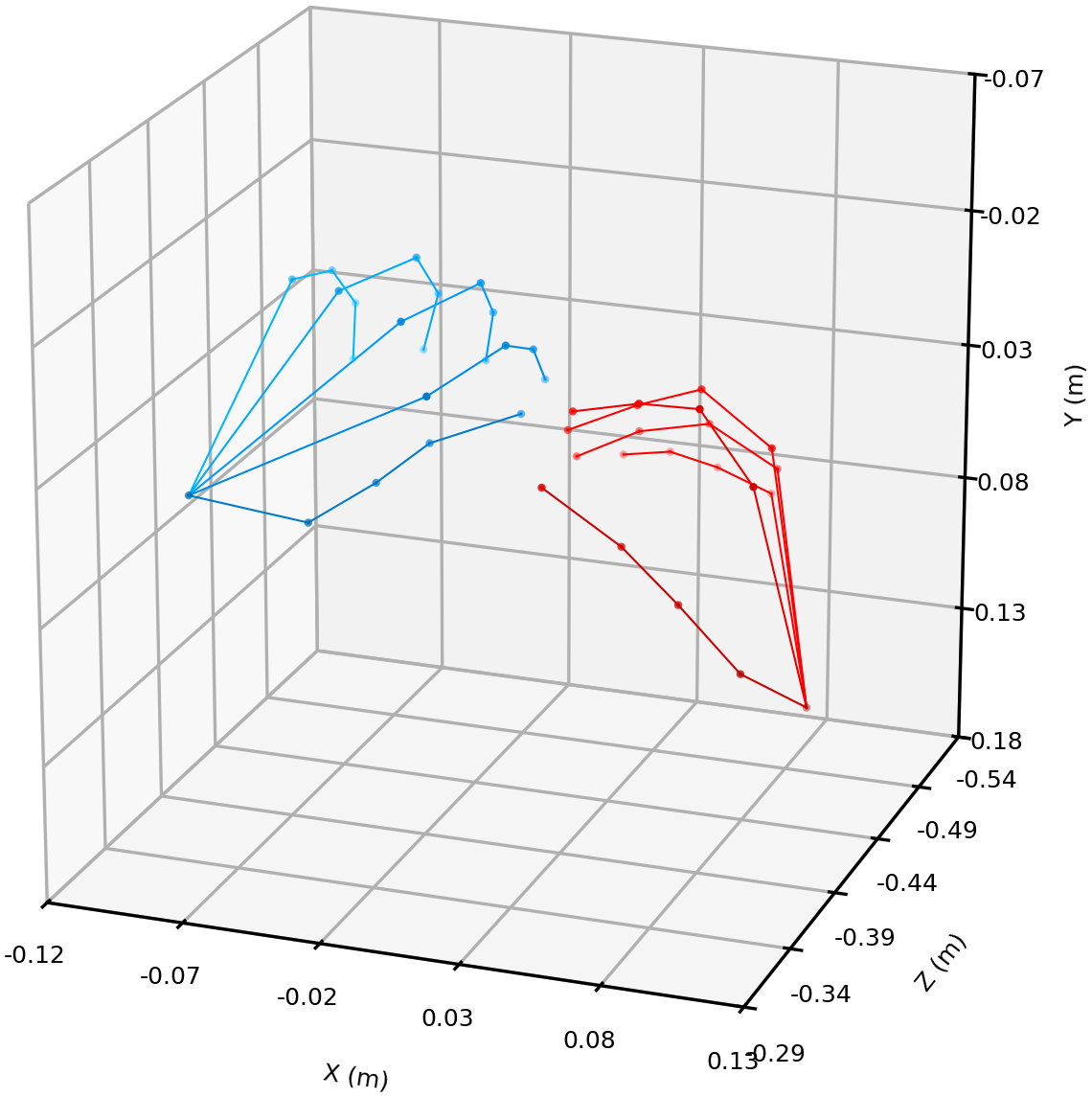} &
\includegraphics[width=\linewidth]{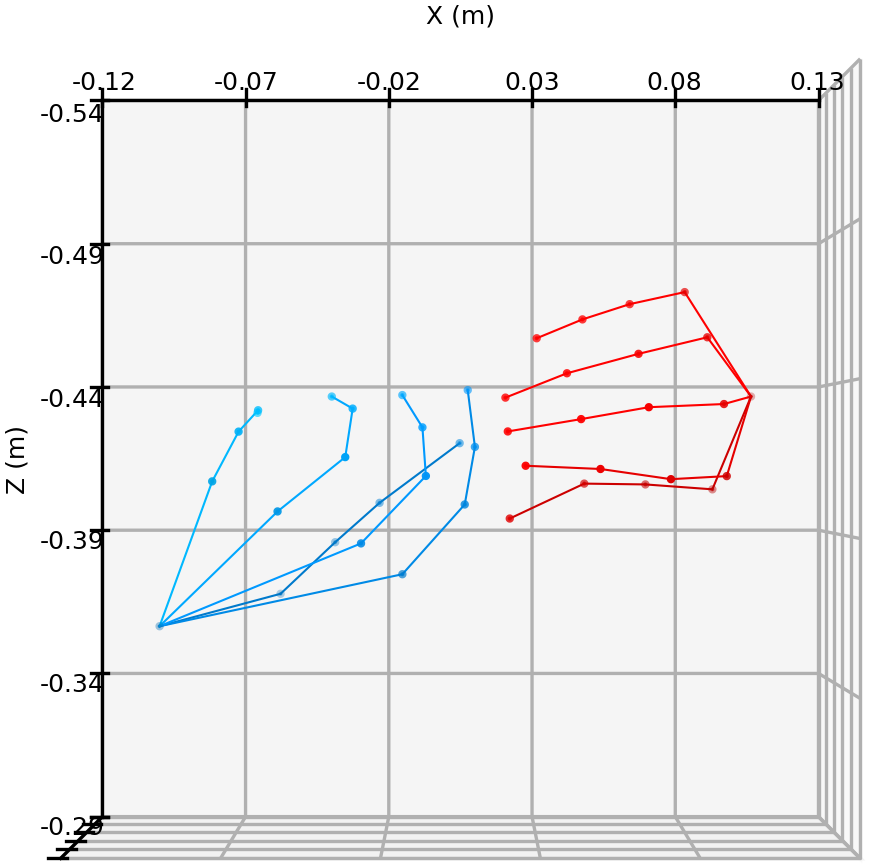}
\end{tabular}
\caption{Qualitative Results of ScaleHP on In-the-Wild Images.
Each panel shows input, box and pose, metric 3D, and top view from left to right.
ScaleHP remains consistent under hand--object and hand--hand interaction,
viewpoint changes, and occlusion; no metric ground truth is available.}
\label{fig:qualitative_wild}
\end{figure*}

\subsection{The Alleviation of the Three Error Sources}

In this subsection, we demonstrate that ScaleHP jointly optimizes the three
coupled error sources: relative geometry, root localization, and their
scale-mediated coupling. All diagnostics use a consistent checkpoint protocol. For
each dataset, CS-MPJPE, aligned MPJPE, axis-wise error, and scale error are
computed from the same predictions of the same dataset-specific ScaleHP
checkpoint. They are distinct from the explicitly cross-dataset protocol in
Table~\ref{tab:cross_dataset}; no aligned and camera-space metrics are selected from
different checkpoints within a dataset.

\subsubsection{Relative Geometry (a).}
\begin{table}[t]
\centering
{\small
\setlength{\tabcolsep}{1pt}
\begin{tabular}{@{}lcccc@{}}
\toprule
\multirow{2}{*}{Method} & FreiHand & \multicolumn{2}{c}{DexYCB} & HO3Dv3 \\
& PA-MPJPE $\downarrow$ & R-MPJPE $\downarrow$ & PA-MPJPE $\downarrow$ &
PA-MPJPE $\downarrow$ \\ \midrule
CMR-PG & 6.9 & - & - & - \\
I2L-MeshNet & 7.4 & - & - & - \\
HandDGP & 7.4 & - & - & - \\
MobRecon & 5.7 & 14.2 & 6.4 & - \\
METRO & 6.7 & 15.2 & 7.0 & - \\
HandOccNet & - & 14.0 & 5.8 & - \\
H2ONet & - & 14.0 & 5.7 & - \\
Deformer & - & 13.6 & 5.2 & - \\
Zhou et al. & - & 12.4 & 5.5 & - \\
TI-Net & - & 16.8 & 4.9 & - \\
MaskHand & 5.5 & 11.7 & 5.0 & 7.0 \\
Hamba & 5.7 & - & - & 6.9 \\
HandOS & \textbf{5.0} & - & 5.2 & 6.8 \\
\textbf{ScaleHP} & \textbf{5.0} & \textbf{10.3} & \textbf{4.6} & \textbf{5.9} \\ \bottomrule
\end{tabular}}
\caption{Aligned/root-relative accuracy (mm). PA-MPJPE applies Procrustes
alignment; R-MPJPE root-aligns translation only.}
\label{tab:aligned_main}
\end{table}

Table~\ref{tab:aligned_main} tests whether absolute metric reasoning compromises
articulation after global factors are removed. ScaleHP ties the best FreiHand
PA-MPJPE (5.0\,mm), reduces DexYCB R-MPJPE and PA-MPJPE to 10.3 and 4.6\,mm, and
reduces HO3Dv3 PA-MPJPE to 5.9\,mm. Thus the camera-space gains in
Tables~\ref{tab:freihand_main} and~\ref{tab:cross_dataset} do not arise by
trading away root-relative geometry.
Because Procrustes alignment removes translation, rotation, and common scale,
these results measure articulated geometry rather than a favorable metric
composition. The result is therefore complementary to CS-MPJPE: scale-aware
learning improves the local representation that is later lifted into camera
space, rather than only correcting its final global gauge.

\label{sec:mitigate}
\begin{table}[t]
\centering
{\small
\setlength{\tabcolsep}{0pt}
\begin{tabular}{@{}>{\raggedright\arraybackslash}p{0.30\columnwidth}
>{\centering\arraybackslash}p{0.175\columnwidth}
>{\centering\arraybackslash}p{0.175\columnwidth}
>{\centering\arraybackslash}p{0.175\columnwidth}
>{\centering\arraybackslash}p{0.175\columnwidth}@{}}
\toprule
\multirow{2}{*}{Setting} & PA-MPJPE & X error & Y error & Z error \\
& $\downarrow$ & $\downarrow$ & $\downarrow$ & $\downarrow$ \\
\midrule
No scale-token & 5.6 & 2.3 & 2.4 & 3.5 \\
\textbf{ScaleHP} & \textbf{5.0} & \textbf{2.2} & \textbf{2.2} & \textbf{3.0} \\
\bottomrule
\end{tabular}}
\caption{Scale-token regularization on FreiHand (mm). The first setting omits
scale-token interaction; its global metric composition uses a fixed dataset
scalar. The X/Y/Z columns are root-relative axis errors from the same prediction
dump as PA-MPJPE.}
\label{tab:scale_token_regularization}
\end{table}

The controlled analysis in Table~\ref{tab:scale_token_regularization} identifies
how the scale token improves that geometry. Removing the token degrades every
relative component,
with the largest change on
the depth axis: Z-axis error increases by 0.5\,mm, versus 0.1 and 0.2\,mm on the
X and Y axes. Because PA-MPJPE removes the global metric gauge, this result cannot
be explained by a better final rescaling. It shows that scale supervision acts as
a regularizer during joint 2D--3D lifting, reducing the relative-depth component
of error source (a). This anisotropic improvement is consequential under
perspective projection: relative-depth error is the component most readily
confounded with normalized root depth during metric composition.

\subsubsection{Global Localization (b).}
Because $\lVert\mathbf t\rVert\gg\lVert\mathbf J\rVert$, root-depth error can
remain visually hidden under accurate reprojection.
Fig.~\ref{fig:qualitative_main} holds image evidence fixed and exposes this
error in metric/top views: all four inputs permit plausible image-plane overlays,
but the green HandDGP and red CMR predictions retain visible root-depth offsets.
The blue ScaleHP roots and joint sets remain closest to yellow GT along the
camera-depth direction. These views therefore diagnose error source (b), which
cannot be inferred from reprojection alone; CS-MPJPE then measures its combined
effect with (a) and (c).

\subsubsection{Scale-Mediated Coupling (c).}
The preceding results establish that the scale token benefits local geometry and
global placement together. Table~\ref{tab:joint_training_accuracy} tests whether this
benefit requires scale to participate while the pose representation is learned.

\begin{table}[t]
\centering
{\small
\setlength{\tabcolsep}{2pt}
\begin{tabular}{@{}lccc@{}}
\toprule
\multirow{2}{*}{Training schedule} & \multirow{2}{*}{PA-MPJPE $\downarrow$} &
\multirow{2}{*}{CS-MPJPE $\downarrow$} & Scale $L_1$ error \\
& & & ($10^{-3}$ m) $\downarrow$ \\ \midrule
Delayed introduction & 5.2 & 39.5 & 2.06 \\
\textbf{Joint} & \textbf{5.0} & \textbf{35.8} & \textbf{1.94}\\ \bottomrule
\end{tabular}}
\caption{Scale-token introduction on FreiHand (mm). Delayed
introduction first pre-trains without the token, then fine-tunes the entire
interactive decoder after inserting it; all token-to-token attention is
enabled. Joint training includes the token from initialization.}
\label{tab:joint_training_accuracy}
\end{table}

Both schedules enable full attention and update all decoder parameters during
fine-tuning; only the time at which scale enters differs. Joint training improves
PA-MPJPE by 0.2\,mm, CS-MPJPE by 3.7\,mm, and scale error by 5.8\%. The agreement
across relative pose, camera-space pose, and scale supports early scale--pose
co-adaptation as a mechanism for reducing error source (c). In particular, the
camera-space gain is not obtained by sacrificing aligned pose, unlike the
relative--global trade-off motivating Sec.~\ref{sec:why_scale}.
The qualitative results make ScaleHP's improved global localization directly
visible; further quantitative comparisons are provided in the supplementary
material.

\subsection{Ablation Study}

\begin{table}[!t]
\centering
{\small
\setlength{\tabcolsep}{0.5pt}
\begin{tabular}{@{}>{\raggedright\arraybackslash}p{0.38\columnwidth}
>{\centering\arraybackslash}p{0.20\columnwidth}
>{\centering\arraybackslash}p{0.20\columnwidth}
>{\centering\arraybackslash}p{0.20\columnwidth}@{}}
\toprule
\multirow{2}{*}{Setting} & FreiHand & DexYCB & HO3Dv3 \\
& CS-MPJPE $\downarrow$ & CS-MPJPE $\downarrow$ & CS-MPJPE $\downarrow$ \\
\midrule
No token + oracle scale & 44.3 & 40.2 & 42.0 \\
\textbf{ScaleHP} & \textbf{35.8} & \textbf{30.0} & \textbf{33.4} \\ \bottomrule
\end{tabular}}
\caption{Per-instance scale-token coupling (mm). Each dataset uses its own
checkpoint; the no-token setting receives an oracle dataset-level global scale.}
\label{tab:scale_token_coupling}
\end{table}

\noindent\textbf{Scale-token coupling.}
Table~\ref{tab:scale_token_coupling} isolates the effect of an instance scale
from a favorable dataset-level calibration. Even after the no-token model receives
an oracle global scale, its CS-MPJPE is higher by 8.5, 10.2, and 8.6\,mm on
FreiHand, DexYCB, and HO3Dv3, respectively. A fixed scalar can correct an average
dataset gauge but cannot represent the scale variation that composes with each
hand's relative geometry and normalized root translation. The result therefore
directly supports the need for per-instance scale in the scale-mediated term.
The global scalar is selected separately on every target test set, making this
an intentionally favorable calibration for the no-token baseline rather than a
deployment setting; the remaining gap is therefore instance-level rather than
dataset-level scale variation.

\noindent\textbf{IMR scale target.}
The predicted scale $L_1$ errors are 1.94, 0.56, and
$1.02\times10^{-3}$\,m on FreiHand, DexYCB, and HO3Dv3. IMR has a higher
between-subject-to-within-subject F-ratio than the thumb/little-finger target
(6.17 versus 2.95) and stronger rank consistency with full-hand size (0.982
versus 0.868), supporting stable scalar supervision. In other words, IMR retains
between-subject size variation while suppressing the finger-specific variation
that makes a single short bone a noisier proxy.

\noindent\textbf{Background intervention.}
White-background FreiHand evaluation changes CS-MPJPE from 35.85 to 37.22\,mm
(3.8\%, within 5\%). Together with scale-token interaction with $\mathbf J^{2D}$
and $\mathbf J^{3D}$, this supports hand-specific evidence rather than a full-image
scene cue. The paired distribution and limitations are reported in the
supplement; this intervention does not establish strict background invariance.

\section{Conclusion}

We introduced ScaleHP, a unified framework for calibrated metric-space hand pose
estimation. Its central contribution is an explicit per-instance metric scale
that interfaces unit-scale local geometry with global positioning. A dedicated
scale token interacts with 2D/3D joint queries, and a parameter-free calibrated
solver recovers normalized translation. ScaleHP achieves state-of-the-art
camera-space accuracy on FreiHand and PA-MPJPE on FreiHand, DexYCB, and HO3Dv3,
showing that global localization and local articulation improve together. Its
transfer results require neither target-dataset training nor oracle scale, while
ablations support instance-conditioned, joint scale--pose learning.

\clearpage
\appendix
\input{supplementary}

\bibliography{ScaleHP}
\end{document}

%% file: supplementary.tex
\section{Detailed Architecture}
\label{sec:supp_architecture}

\subsection{DETR-Like Detector}

ScaleHP uses a frozen Transformer-based detector to localize hands from a text
prompt. We adopt Grounding DINO \citep{Liu2023GroundingDM} as the representative
implementation (Grounding DINO 1.5 in our experiments), while the framework is
compatible with DETR-like or open-vocabulary detectors that provide multimodal
features and query-based localization. Given an RGB image $\mathbf I$ and the
generic prompt \textit{"hand"}, the encoder $\mathcal E$ fuses the outputs of the
visual and text backbones:
\begin{equation}
\mathbf F_e=\mathcal E\!\left(
\operatorname{Backbone}_v(\mathbf I),
\operatorname{Backbone}_t(\text{"hand"})\right).
\label{eq:supp_detector_encoder}
\end{equation}
The Transformer decoder $\mathcal D$ iteratively refines content queries
$\mathbf Q_i$ and their 2D reference points $\mathbf R_i$. Deformable attention
\citep{DBLP:conf/iclr/ZhuSLLWD21} lets each query sample the encoded multi-scale
features around its reference point and thereby capture local hand context:
\begin{equation}
(\mathbf Q_{i+1},\mathbf R_{i+1})=
\mathcal D_i(\mathbf Q_i,\mathbf R_i,\mathbf F_e).
\label{eq:supp_detector_decoder}
\end{equation}
This process produces multi-scale image features $\mathbf F_{GD}$; the refined
$\mathbf Q$ encodes holistic hand semantics, while $\mathbf R$ carries spatial
locations. The detection head $\mathcal H$ maps them to bounding boxes
$\mathbf B$ and confidence scores $\mathbf C$:
\begin{equation}
(\mathbf B,\mathbf C)=\mathcal H(\mathbf Q,\mathbf R).
\label{eq:supp_detector_head}
\end{equation}
The resulting detector representation is
\begin{equation}
\left\{\mathbf F_{GD},\mathbf Q,\mathbf R,\mathbf B,\mathbf C\right\},
\label{eq:supp_detector_output}
\end{equation}
which is passed to the Metric 2D-3D Decoder. After score filtering and NMS,
suppose $K$ hands remain, represented by
$\{(\tilde{\mathbf Q}_k,\tilde{\mathbf R}_k,\tilde{\mathbf B}_k)\}_{k=1}^{K}$.
Each hand has its own detector query and reference point, and the same hand
decoder processes all $K$ retained instances in parallel. ScaleHP therefore
handles multiple hands without changing the decoder formulation. The detector
remains frozen, so the scale, pose, and projection losses optimize only the hand
decoder. The full image is resized to a 1280-pixel long edge and no externally
supplied hand crop is required.

\subsection{Metric 2D-3D Decoder}

\noindent\textbf{Instance expansion.}
For each retained hand, learnable joint embeddings $\mathbf E_J$ expand the
instance query into two semantically indexed sets of $J=21$ tokens. The scale
token shares the same instance context but has its own stream:
\begin{equation}
\mathbf Q^{2D}_0=\mathbf Q^{3D}_0=\tilde{\mathbf Q}+\mathbf E_J,
\qquad \mathbf q^s_0=\tilde{\mathbf Q}.
\label{eq:supp_query_initialization}
\end{equation}
The 2D tokens represent joint locations in the image, the 3D tokens represent
the corresponding unit-scale root-relative joints, and the single scale token
represents a hand-level metric gauge. Initial spatial references are inherited
from the corresponding detector instance and subsequently refined by each stream.

\noindent\textbf{Reciprocal scale-joint interaction.}
Let $\mathbf Q^J_l=[\mathbf Q^{2D}_l;\mathbf Q^{3D}_l]$ collect the two joint
streams. At interactive layer $l$, the scale and joint streams are concatenated
without a directional attention mask:
\begin{equation}
\mathbf X_{l-1}=[\mathbf q^s_{l-1};\mathbf Q^J_{l-1}],\qquad
[\bar{\mathbf q}^s_l;\bar{\mathbf Q}^J_l]=\operatorname{SA}_l(\mathbf X_{l-1}).
\label{eq:supp_reciprocal_attention}
\end{equation}
Consequently, the scale-token output aggregates keys and values from named 2D
and 3D joints, while each joint output contains an attention-weighted
contribution from the scale token. The former direction lets scale estimation
use joint layout and normalized skeletal structure; the latter makes
scale-supervised context available to pose features. Since the resulting states
are fed to the next layer, repeated interaction permits reciprocal refinement
rather than a one-time concatenation at the prediction head.

\noindent\textbf{Image interaction and spatial refinement.}
Self-attention is followed by stream-specific deformable cross-attention:
\begin{equation}
\begin{gathered}
\mathbf Q^d_l=\operatorname{DefAttn}^d_l
(\bar{\mathbf Q}^d_l,\mathbf F_{GD},\mathbf R^d_{l-1}),\\
\mathbf R^d_l=\mathbf R^d_{l-1}+
\operatorname{FFN}^d_R(\mathbf Q^d_l),\qquad d\in\{s,2D,3D\}.
\end{gathered}
\label{eq:supp_deformable_update}
\end{equation}
where $\mathbf Q^s_l\equiv\mathbf q^s_l$. Thus the scale token combines
skeletal evidence with multi-scale hand appearance and image occupancy, while
2D/3D tokens sample local evidence near their evolving references. Three
independent heads at the final layer output
\begin{equation}
\begin{gathered}
\mathbf J^{2D}=\operatorname{FFN}_{2D}(\mathbf Q^{2D}_L),\qquad
\mathbf J^{3D}=\operatorname{FFN}_{3D}(\mathbf Q^{3D}_L),\\
s=\operatorname{FFN}_{s}(\mathbf q^s_L).
\end{gathered}
\label{eq:supp_decoder_outputs}
\end{equation}
They are supervised by pixel joints $\mathbf J^{2D*}$, unit-scale joints
$\mathbf J^{3D*}$, and IMR scale $s^*$ derived from the same metric annotation
$\mathbf J^{M*}$. The analytical transfer below consumes these three outputs
and known intrinsics; no learned translation head or iterative registration is
used.

\section{Derivation of Analytical Metric Transfer}
\label{sec:supp_transfer}

Once the global scale $s$ is predicted, normalized root translation can be
recovered from perspective constraints. Let joint $j$ have predicted pixel
coordinates $(u_j,v_j)$ and unit-scale root-relative coordinates
$(x_j,y_j,z_j)$. We first derive the $x$-axis relation:
\begin{equation}
\frac{u_j-c_x}{f_x}
=\frac{x_j+t_x^{norm}}{z_j+t_z^{norm}}.
\label{eq:supp_projection}
\end{equation}
Multiplying both sides by the denominators gives
\begin{equation}
(u_j-c_x)(z_j+t_z^{norm})
=f_x(x_j+t_x^{norm}).
\label{eq:supp_projection_expanded}
\end{equation}
Let $p$ be the root joint, which is the origin of the root-relative coordinate
system. Setting $(x_p,y_p,z_p)=\mathbf 0$ in
Eq.~\ref{eq:supp_projection_expanded} yields
\begin{equation}
(u_p-c_x)t_z^{norm}=f_x t_x^{norm}.
\label{eq:supp_root_x}
\end{equation}
Substituting Eq.~\ref{eq:supp_root_x} back into the expanded projection gives
\begin{equation}
(u_j-c_x)(z_j+t_z^{norm})
=f_x\left(x_j+\frac{u_p-c_x}{f_x}t_z^{norm}\right).
\label{eq:supp_projection_substitution}
\end{equation}
Collecting all terms that contain $t_z^{norm}$ on the left produces
\begin{equation}
(u_j-u_p)t_z^{norm}-f_x\,x_j=-(u_j-c_x)z_j.
\label{eq:supp_projection_rearranged}
\end{equation}
The same steps along the $y$-axis give
\begin{equation}
(v_j-v_p)t_z^{norm}-f_y\,y_j=-(v_j-c_y)z_j.
\label{eq:supp_projection_rearranged_y}
\end{equation}
Stacking both equations for every non-root joint forms the overdetermined
linear system
\begin{equation}
\underbrace{\begin{bmatrix}
\vdots\\ u_j-u_p\\ v_j-v_p\\ \vdots
\end{bmatrix}}_{\mathbf A}t_z^{norm}
=
\underbrace{\begin{bmatrix}
\vdots\\ f_x\,x_j-(u_j-c_x)z_j\\
f_y\,y_j-(v_j-c_y)z_j\\ \vdots
\end{bmatrix}}_{\mathbf b}.
\label{eq:supp_linear_system}
\end{equation}
We solve $t_z^{norm}=\mathbf A^{\dagger}\mathbf b$ by least squares and recover
the other two components from the root projection:
\begin{equation}
t_x^{norm}=\frac{u_p-c_x}{f_x}t_z^{norm},\qquad
t_y^{norm}=\frac{v_p-c_y}{f_y}t_z^{norm}.
\label{eq:supp_root_xy}
\end{equation}
Because a common scale cancels under projection, the system estimates normalized
translation. The predicted scale converts both the root-relative pose and its
translation to metric camera coordinates:
\begin{equation}
\mathbf Q^{cam}=s(\mathbf J^{rel}+\mathbf t^{norm}),\qquad
\mathbf t^M=s\mathbf t^{norm}.
\label{eq:supp_metric_composition}
\end{equation}

\section{Datasets, Training, and Reproducibility Details}
\label{sec:supp_dataset}

\noindent\textbf{Datasets.}
We evaluate the performance of our method on the following three primary benchmarks, which provide standardized protocols for 3D hand pose estimation:

\textbf{FreiHand} \citep{zimmermann2019freihand}: A widely recognized benchmark for 3D hand pose and MANO \citep{Romero2017EmbodiedH} fitting, containing 130,240 training and 3,960 test images. We follow the evaluation protocols in \citep{DBLP:conf/cvpr/ChenLMCWCGWZ21, Huang_2023_CVPR}, utilizing the provided 3D joint annotations for both supervision and quantitative assessment.

\textbf{DexYCB} \citep{chao2021dexycb}: A large-scale benchmark for 3D hand pose estimation and hand-object interaction. It comprises 429,616 training and 78,768 testing samples, featuring synchronized RGB-D sequences of human hands interacting with 20 YCB objects. The dataset provides accurate 3D annotations for both hand joints and object poses, offering an ideal platform for evaluating models in complex, real-world grasping scenarios with significant occlusions.

\textbf{HO3Dv3} \citep{Hampali2019HOnnotateAM}: This refined version of the HO3D dataset offers enhanced annotation accuracy and a larger scale, comprising 83,325 training and 20,137 test images. The dataset captures real-world 3D hand-object interactions and presents a significant challenge due to the severe occlusions caused by the manipulated objects.

\label{sec:supp_reproducibility}

\noindent\textbf{Optimization.}
The frozen Grounding DINO 1.5 detector processes the full image after resizing
its long edge to 1280 pixels. The hand Decoder is trained for 45 epochs with
Adam, total batch size 16, initial learning rate $10^{-4}$, and eight NVIDIA
A100-80GB GPUs. The recorded configuration uses random seed 21 and a
cosine-annealing schedule from epoch 7 to 45 with minimum learning rate
$5\times10^{-6}$. The Decoder has six interactive loss layers, 21 joints per
hand, and four image-feature levels. The first two layers are 2D-only; scale,
projection, and 3D supervision begin at layer 2.

\noindent\textbf{Training objective.}
Let $\ell\in\{0,\ldots,5\}$ index the six decoder outputs. The complete enabled
objective and its configured weights are
\begin{equation}
\begin{aligned}
\mathcal L={}&\sum_{\ell=0}^{5}\left(
10\mathcal L^{2D}_{1,\ell}
+4\mathcal L^{2D}_{\mathrm{OKS},\ell}\right)\\
&+\sum_{\ell=2}^{5}\left(
20\mathcal L^{s}_{\mathrm{MSE},\ell}
+10\mathcal L^{\mathrm{proj}}_{1,\ell}
+4\mathcal L^{\mathrm{proj}}_{\mathrm{OKS},\ell}\right.\\
&\hspace{32mm}\left.
+10\mathcal L^{3D}_{1,\ell}
+5\mathcal L^{\mathrm{edge}}_{\ell}\right).
\end{aligned}
\label{eq:supp_training_loss}
\end{equation}
Thus the shared $L_1$, OKS, relative-edge, and scale-MSE weights are
$10.0$, $4.0$, $5.0$, and $20.0$, respectively. The projection terms reuse the
same $L_1$/OKS weights.

\section{Accuracy of Metric Scale Prediction}

To evaluate the learned per-instance metric gauge separately from pose and
translation, we report its average $L_1$ error against the annotation-derived
target on three datasets. Table~\ref{tab:scale_acc} shows errors below
$2\times10^{-3}$ meters. This result measures in-distribution prediction of the
supervised scale target.

\begin{table}[t]
\centering
\setlength{\tabcolsep}{6pt}
\begin{tabular}{lccc}
\toprule
Method
& FreiHand
& DexYCB
& HO3Dv3 \\
\midrule
\textbf{Ours}
& \textbf{1.94}
& \textbf{0.56}
& \textbf{1.02} \\
\bottomrule
\end{tabular}
\caption{Per-instance scale prediction error across datasets ($L_1$ distance, $10^{-3}$ meters).}
\label{tab:scale_acc}
\end{table}

\section{Camera-Space Evaluation Protocol}

Camera-space comparisons depend on the information supplied at test time.
ScaleHP consumes RGB and known camera intrinsics, but no ground-truth hand
scale. The 42.4\,mm NVF result in the main paper uses FreiHand's reference-bone
scale; NVF reports 47.2\,mm without it. For CMR and HandDGP, ``+GS'' denotes a
test-set-optimal dataset-level rescaling applied to root-relative geometry and
translation. This oracle calibration is used only for the cross-dataset control.

\noindent\textbf{Main-paper checkpoint mapping.}
Main-paper Table~2 evaluates a FreiHand-trained checkpoint on FreiHand.
Main-paper Table~3 applies that same checkpoint directly to DexYCB and HO3Dv3,
without target-dataset fine-tuning. In main-paper Table~4, each dataset column
uses a checkpoint trained on that dataset. Main-paper Tables~5 and~6 contain
FreiHand-trained ablations, whereas each column of main-paper Table~7 uses its
corresponding dataset-specific checkpoint. The supplementary root-localization,
white-background, and FreiHand qualitative results use the FreiHand checkpoint;
the scale-token interaction ablation uses the corresponding checkpoint trained
on each of the three datasets. All reported values are single-checkpoint point
estimates computed consistently from the same prediction dump for each setting.

\section{Global Root Localization Error}
\label{sec:supp_root_error}

To verify that joint optimization effectively reduces error source~(b), we
compare predicted wrist-root positions with ground truth in the FreiHand camera
coordinate system. Table~\ref{tab:supp_root_official} reports the mean
Euclidean root distance together with mean absolute errors on the three camera
axes. For CMR, the reported 48.8\,mm result belongs to CMR-PG with an unreleased
ResNet50 backbone; our reproducible evaluation therefore uses the public CMR-SG
ResNet18 checkpoint, its native 224$\times$224 input and $0.2$ scale, and the
authors' UV/confidence/silhouette registration. HandDGP uses its public
FreiHand checkpoint, rectified 224$\times$224 input, canonical focal length
1000, principal point 112, and native $0.2$ hand scale; this setting reproduces
the paper result at 46.35\,mm versus the reported 46.3\,mm.

\begin{table}[t]
\centering
{\footnotesize
\setlength{\tabcolsep}{3.3pt}
\begin{tabular}{@{}lrrrr@{}}
\toprule
Method & Root L2 $\downarrow$
& $|\Delta X|\downarrow$ & $|\Delta Y|\downarrow$ & $|\Delta Z|\downarrow$ \\ \midrule
CMR & 50.14 & 5.34 & 4.96 & 48.59 \\
HandDGP & 46.19 & 5.33 & 4.34 & 44.65 \\
\textbf{ScaleHP} & \textbf{35.74} & \textbf{4.34} & \textbf{3.73} & \textbf{34.36} \\ \bottomrule
\end{tabular}}
\caption{FreiHand root-localization errors (mm). The best CMR-PG result in the
paper is 48.8\,mm CS-MPJPE, but its ResNet50 weights were not released; the CMR
row therefore uses the public CMR-SG checkpoint.}
\label{tab:supp_root_official}
\end{table}

As shown in Table~\ref{tab:supp_root_official}, ScaleHP reduces mean root error
by 10.45\,mm relative to HandDGP and by 14.40\,mm relative to the reproduced
CMR model. The corresponding depth-error reductions are 10.29 and 14.23\,mm,
whereas the lateral-axis changes are below 1.4\,mm. Thus the gain is concentrated
on camera-depth localization, demonstrating that the joint scale-pose
optimization directly suppresses the dominant component of error source~(b).

\FloatBarrier
\section{Scale-Token-Joint Interaction}
\label{sec:supp_controls}

Table~\ref{tab:supp_scale_interaction} isolates how the metric scale
representation communicates with joint queries while keeping the detector,
data, objective, and optimization schedule fixed. The oracle-global-scale
control removes per-instance scale prediction and uses the test-set-optimal
dataset-level scalar. In the independent-branch variant, the filtered instance
query is mapped directly to $s$ by an MLP with scale supervision, without
scale-joint attention. The two one-way variants
apply a self-attention mask: either the scale token can attend to joint queries
but joints cannot attend back, or joints can attend to scale but scale cannot
attend back. The proposed model leaves both directions open.

\begin{center}
\centering
{\small
\setlength{\tabcolsep}{4pt}
\begin{tabular}{@{}lccc@{}}
\toprule
Setting & FreiHand $\downarrow$ & DexYCB $\downarrow$ & HO3Dv3 $\downarrow$ \\ \midrule
Oracle global scale & 44.3 & 40.2 & 42.0 \\
Independent branch & 42.9 & 37.8 & 40.5 \\
Scale sees joints only & 40.4 & 35.3 & 38.0 \\
Joints see scale only & 40.2 & 35.4 & 37.6 \\
\textbf{Bidirectional interaction} & \textbf{35.8} & \textbf{30.0} & \textbf{33.4} \\ \bottomrule
\end{tabular}}
\captionof{table}{Scale-token interaction ablation measured by CS-MPJPE (mm). Lower is
better. Each dataset uses its corresponding within-dataset checkpoint.}
\label{tab:supp_scale_interaction}
\end{center}

Both one-way interactions improve over an independent scale branch, but neither
matches reciprocal attention. As shown in
Table~\ref{tab:supp_scale_interaction}, bidirectional exchange lowers CS-MPJPE
by between 4.4 and 5.4\,mm relative to the stronger one-way result on each
dataset. This
consistent gain shows that the two directions solve complementary problems:
joint geometry improves scale estimation, while scale context improves joint
representation. Reciprocal interaction, rather than merely adding a scale
branch, is therefore responsible for the best camera-space accuracy.

\section{Aligned-Pose Comparison}

Table~\ref{tab:supp_aligned} evaluates articulated geometry after removing
global factors. ScaleHP ties the best FreiHand Procrustes-aligned error and
achieves the lowest listed errors on DexYCB and HO3Dv3. These results show that
the camera-space improvement is not obtained by sacrificing root-relative hand
geometry.

\begin{table*}[!t]
\centering
{\footnotesize
\renewcommand{\arraystretch}{0.92}
\setlength{\tabcolsep}{4pt}
\begin{tabular}{l|cc|cc|c}
\toprule
Method
& \multicolumn{2}{c|}{FreiHand}
& \multicolumn{2}{c|}{DexYCB}
& HO3Dv3 \\
\cmidrule(lr){2-3}\cmidrule(lr){4-5}\cmidrule(lr){6-6}
& CS-MPJPE $\downarrow$ & PA-MPJPE $\downarrow$
& R-MPJPE $\downarrow$ & PA-MPJPE $\downarrow$
& PA-MPJPE $\downarrow$ \\ \midrule
ObMan \citep{hasson19_obman} & 85.2 & 13.3 & - & - & - \\
MANO CNN \citep{zimmermann2019freihand} & 71.3 & 11.0 & - & - & - \\
CMR-PG \citep{DBLP:conf/cvpr/ChenLMCWCGWZ21} & 48.8 & 6.9 & - & - & - \\
I2L-MeshNet \citep{moon2020i2l} & 60.3 & 7.4 & - & - & - \\
HandDGP \citep{valassakis2024handdgp} & \underline{46.3} & 7.4 & - & - & - \\
MobRecon \citep{DBLP:conf/cvpr/ChenLDZMXZG22} & 50.2 & 5.7 & 14.2 & 6.4 & - \\
METRO \citep{DBLP:conf/cvpr/LinWL21} & - & 6.7 & 15.2 & 7.0 & - \\
HandOccNet \citep{DBLP:conf/cvpr/ParkOMCL22} & - & - & 14.0 & 5.8 & - \\
H2ONet \citep{Xu2023H2ONetHN} & - & - & 14.0 & 5.7 & - \\
Deformer \citep{Yoshiyasu_2023_CVPR} & - & - & 13.6 & 5.2 & - \\
Zhou et al. \citep{Zhou_2024_CVPR} & - & - & 12.4 & 5.5 & - \\
TI-Net \citep{ren2025learning} & - & - & 16.8 & \underline{4.9} & - \\
MaskHand \citep{Saleem2024MaskHandGM} & - & 5.5 & \underline{11.7} & 5.0 & 7.0 \\
S$^2$HAND \citep{Chen2021Modelbased3H} & - & 11.8 & - & - & 11.5 \\
ArtiBoost \citep{Li2021ArtiBoostBA} & - & - & 12.8 & - & 10.8 \\
HandGCAT \citep{Wang2023HandGCATO3} & - & - & - & - & 9.1 \\
AMVUR \citep{Jiang2023APA} & - & 6.2 & - & - & 8.7 \\
SPMHand \citep{Lu2024SPMHandSP} & - & - & - & - & 8.6 \\
Hamba \citep{DBLP:conf/nips/DongCG0T24} & - & 5.7 & - & - & 6.9 \\
HandOS \citep{chen2025handos} & - & \textbf{5.0} & - & 5.2 & \underline{6.8} \\
\textbf{ScaleHP} & \textbf{35.8} & \textbf{5.0} & \textbf{10.3} & \textbf{4.6} & \textbf{5.9} \\ \bottomrule
\end{tabular}}
\caption{State-of-the-art comparison on FreiHand, DexYCB, and HO3Dv3 (mm).}
\label{tab:supp_aligned}
\end{table*}

\FloatBarrier

\section{White-Background Intervention}

To test whether scale prediction relies on scene background, we replace every
FreiHand evaluation background with white while preserving foreground hand
pixels and their image coordinates, as illustrated in
Fig.~\ref{fig:supp_background_examples}. We then rerun the complete
detector-decoder pipeline on all 3,960 paired samples.

\begin{table}[t]
\centering
{\small
\setlength{\tabcolsep}{5pt}
\begin{tabular}{@{}lr@{}}
\toprule
Statistic & Value \\ \midrule
Original CS-MPJPE & 35.854 mm \\
White-background CS-MPJPE & 37.225 mm \\
Mean paired change & +1.370 mm \\
Median paired change & +0.113 mm \\
Scale Pearson / Spearman & 0.622 / 0.595 \\
Mean detection-box IoU & 0.840 \\
Mean 2D-keypoint displacement & 4.343 px \\ \bottomrule
\end{tabular}}
\caption{Full-pipeline white-background intervention on FreiHand.}
\label{tab:supp_background}
\end{table}

\begin{center}
\centering
\setlength{\tabcolsep}{1pt}
\begin{tabular}{*{4}{>{\centering\arraybackslash}m{0.23\columnwidth}}}
\textbf{Original} & \textbf{White} & \textbf{Original} & \textbf{White} \\ [2pt]
\includegraphics[width=\linewidth]{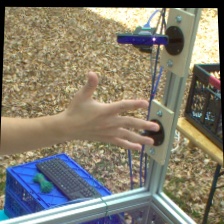} &
\includegraphics[width=\linewidth]{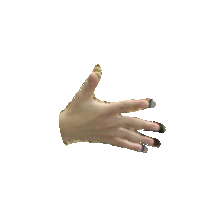} &
\includegraphics[width=\linewidth]{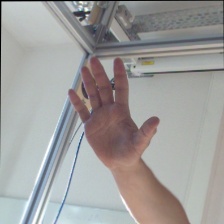} &
\includegraphics[width=\linewidth]{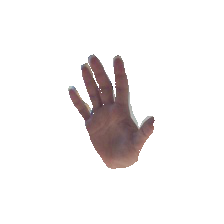} \\[2pt]
\includegraphics[width=\linewidth]{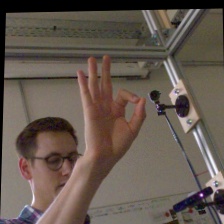} &
\includegraphics[width=\linewidth]{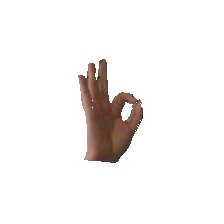} &
\includegraphics[width=\linewidth]{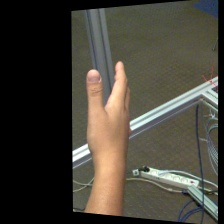} &
\includegraphics[width=\linewidth]{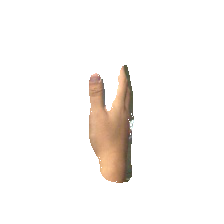}
\end{tabular}
\captionof{figure}{White-background intervention examples. Foreground hand pixels and
their image coordinates are retained, while all other pixels are set to white;
images are shown without cropping.}
\label{fig:supp_background_examples}
\end{center}

Table~\ref{tab:supp_background} shows that removing all background content
changes camera-space error by only 1.370\,mm (3.8\%), while the median paired
change is 0.113\,mm. This small shift persists even though the upstream detector
boxes and 2D joints also change. Together with the retained scale correlation,
the result shows that the scale token is driven primarily by foreground hand
appearance and joint geometry rather than background scene cues.

\FloatBarrier
\section{Additional FreiHand Qualitative Results}

Figures~\ref{fig:supp_freihand_1} and~\ref{fig:supp_freihand_2} show all 16
additional FreiHand groups. Each row shows the input, metric camera-space
reconstruction, and front, side, and top views.

\onecolumn
\raggedbottom
\newcommand{\freihandrow}[1]{%
\includegraphics[width=\linewidth,height=0.107\textheight,keepaspectratio]{optimized_freihand/sample_#1_original.jpg} &
\includegraphics[width=\linewidth,height=0.107\textheight,keepaspectratio]{sample_#1_3d_metric.png} &
\includegraphics[width=\linewidth,height=0.107\textheight,keepaspectratio]{sample_#1_3d_front.png} &
\includegraphics[width=\linewidth,height=0.107\textheight,keepaspectratio]{sample_#1_3d_side.png} &
\includegraphics[width=\linewidth,height=0.107\textheight,keepaspectratio]{sample_#1_3d_top.png} \\[-1pt]}

\begin{center}
\centering
\setlength{\tabcolsep}{0.5pt}
\begin{tabular}{*{5}{>{\centering\arraybackslash}m{0.145\textwidth}}}
\textbf{Input} & \textbf{Camera space} & \textbf{Front} &
\textbf{Side} & \textbf{Top} \\[2pt]
\freihandrow{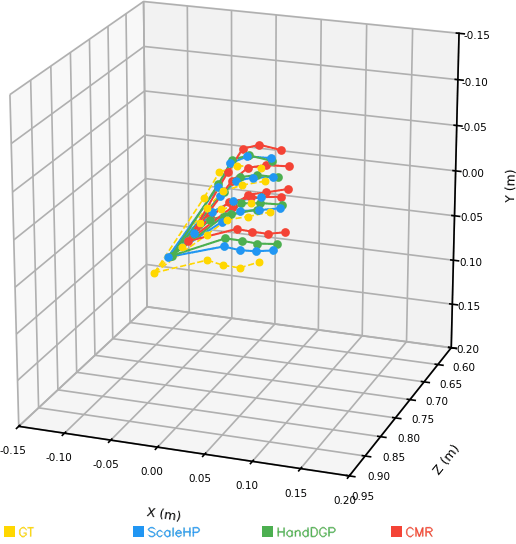}
\freihandrow{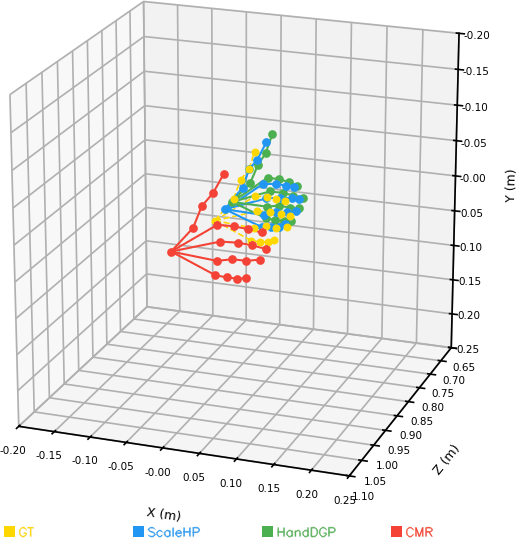}
\freihandrow{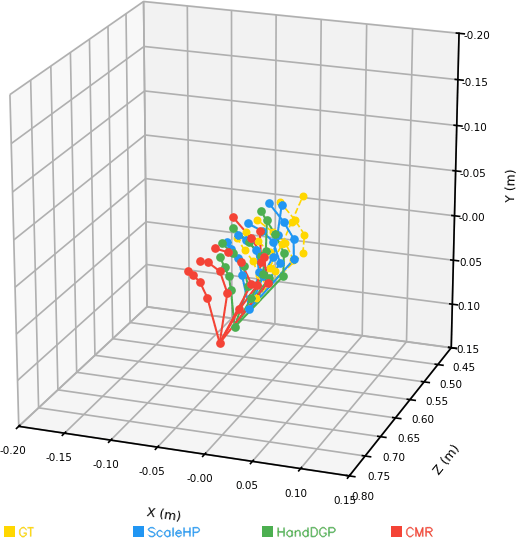}
\freihandrow{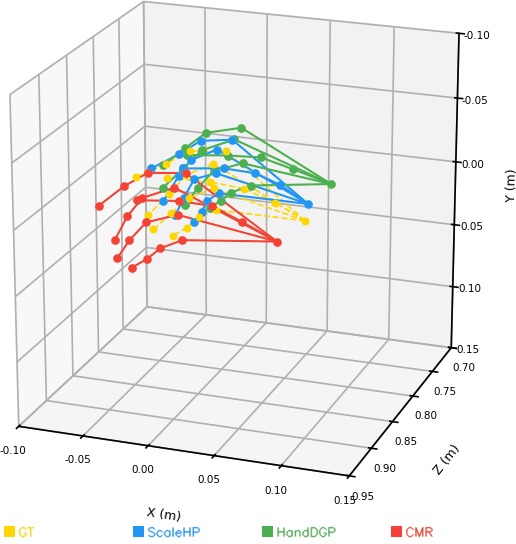}
\freihandrow{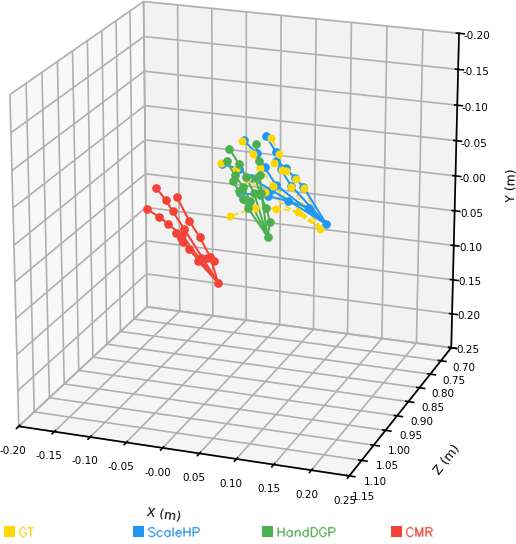}
\freihandrow{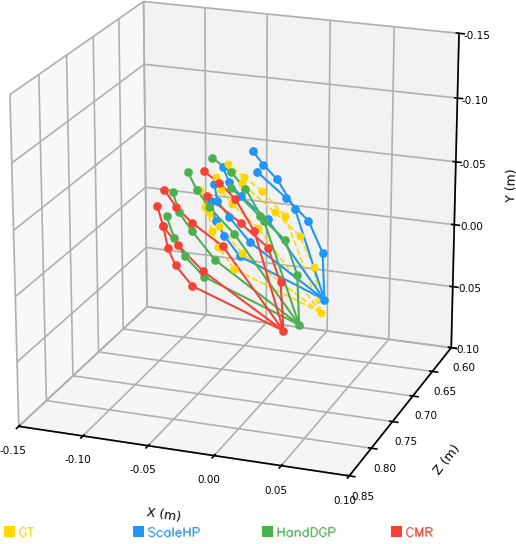}
\freihandrow{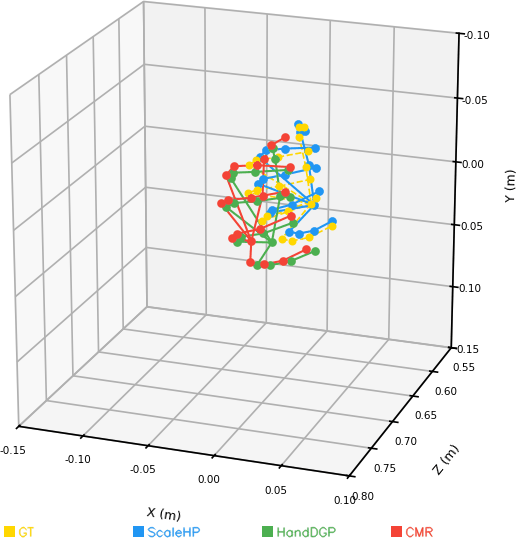}
\freihandrow{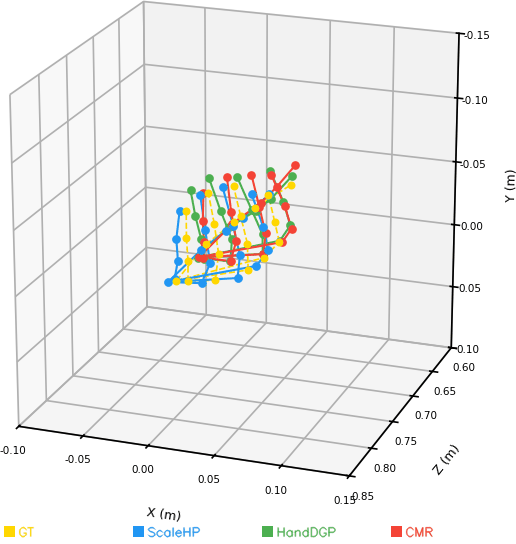}
\end{tabular}
\captionof{figure}{Additional FreiHand results (1/2). Yellow: GT; blue: ScaleHP;
green: HandDGP; red: CMR. Best viewed zoomed in.}
\label{fig:supp_freihand_1}
\end{center}

\clearpage
\begin{center}
\centering
\setlength{\tabcolsep}{0.5pt}
\begin{tabular}{*{5}{>{\centering\arraybackslash}m{0.145\textwidth}}}
\textbf{Input} & \textbf{Camera space} & \textbf{Front} &
\textbf{Side} & \textbf{Top} \\[2pt]
\freihandrow{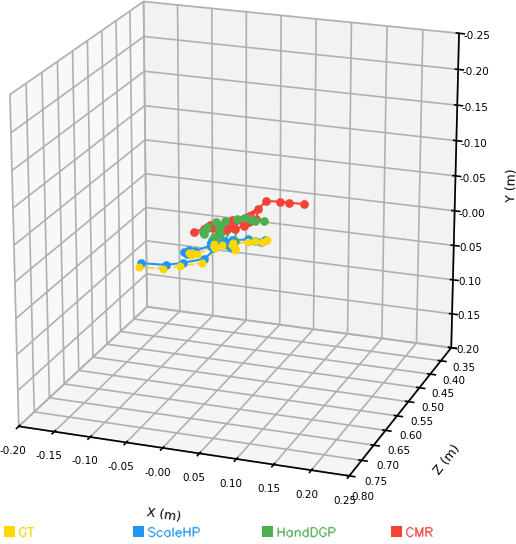}
\freihandrow{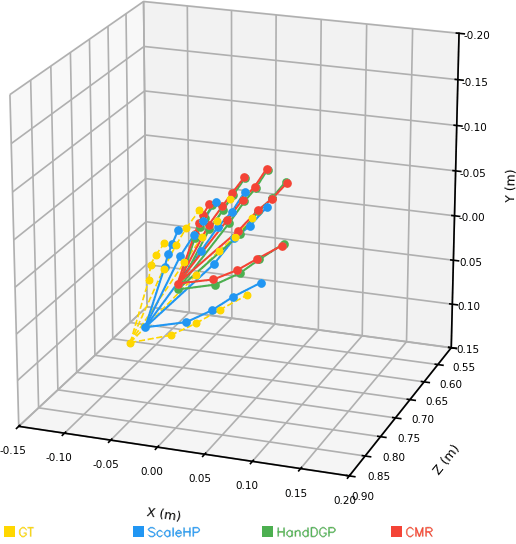}
\freihandrow{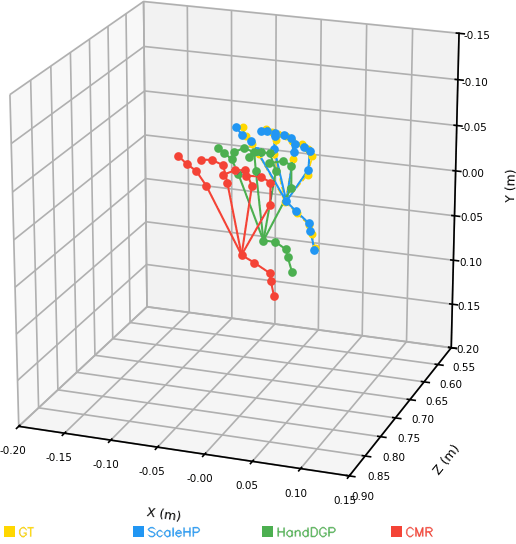}
\freihandrow{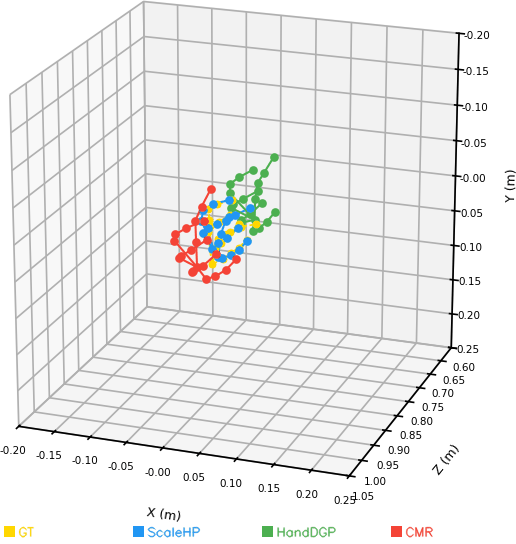}
\freihandrow{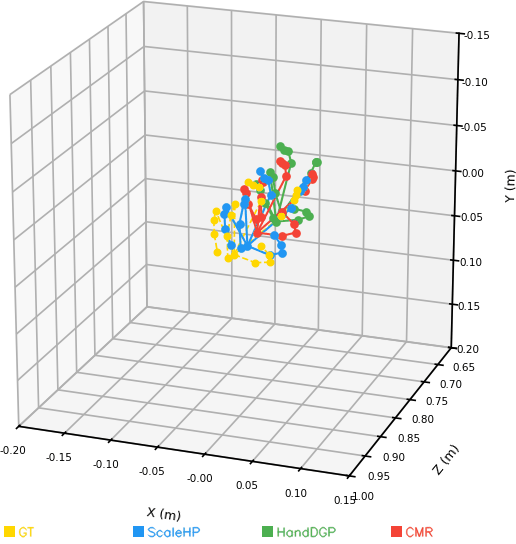}
\freihandrow{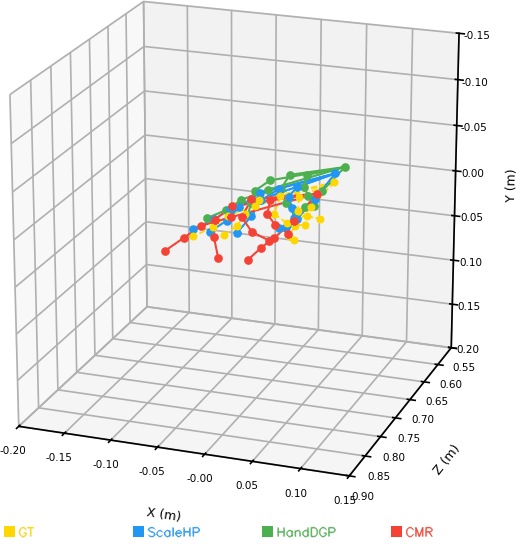}
\freihandrow{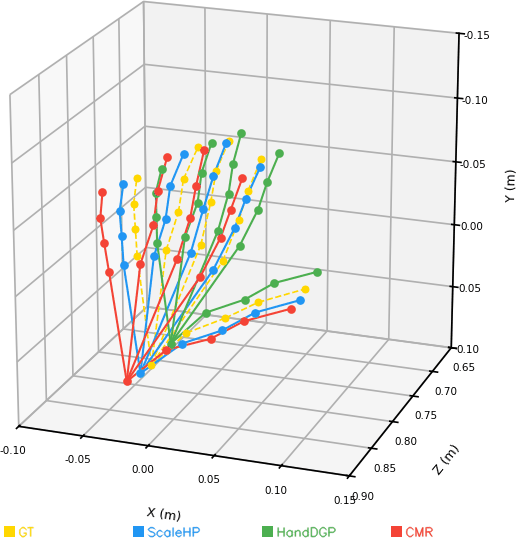}
\freihandrow{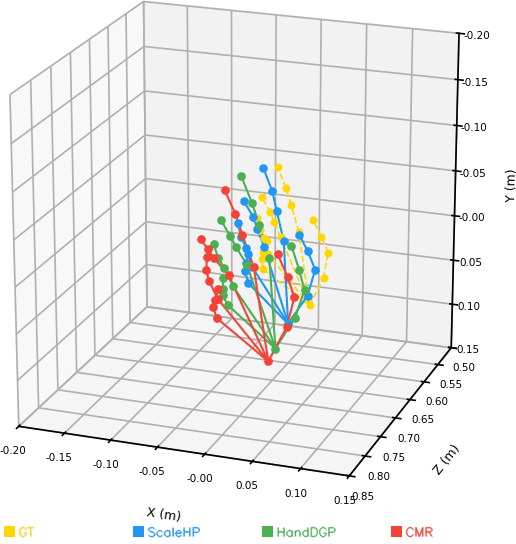}
\end{tabular}
\captionof{figure}{Additional FreiHand results (2/2). Yellow: GT; blue: ScaleHP;
green: HandDGP; red: CMR. Best viewed zoomed in.}
\label{fig:supp_freihand_2}
\end{center}

\clearpage
\section{More ScaleHP Qualitative Results}

Figures~\ref{fig:qualitative_results_1} and~\ref{fig:qualitative_results_2}
show ScaleHP outputs on eight open-domain images. The examples include
occlusion, motion blur, varied lighting, and interaction patterns.

\begin{center}
    \centering
    \setlength{\tabcolsep}{1pt}
    \begin{tabular}{
        >{\centering\arraybackslash}m{0.025\textwidth}
        *{4}{>{\centering\arraybackslash}m{0.225\textwidth}}
    }

     & \textbf{Ex.1} & \textbf{Ex.2} & \textbf{Ex.3} & \textbf{Ex.4} \\[3pt]

    \rotatebox{90}{Input} &
    \includegraphics[width=\linewidth,height=0.125\textheight,keepaspectratio]{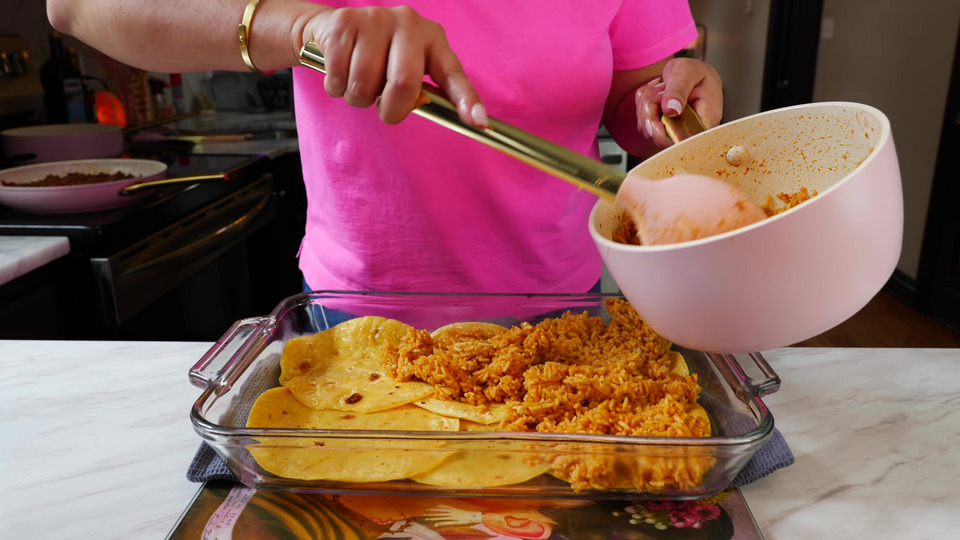} &
    \includegraphics[width=\linewidth,height=0.125\textheight,keepaspectratio]{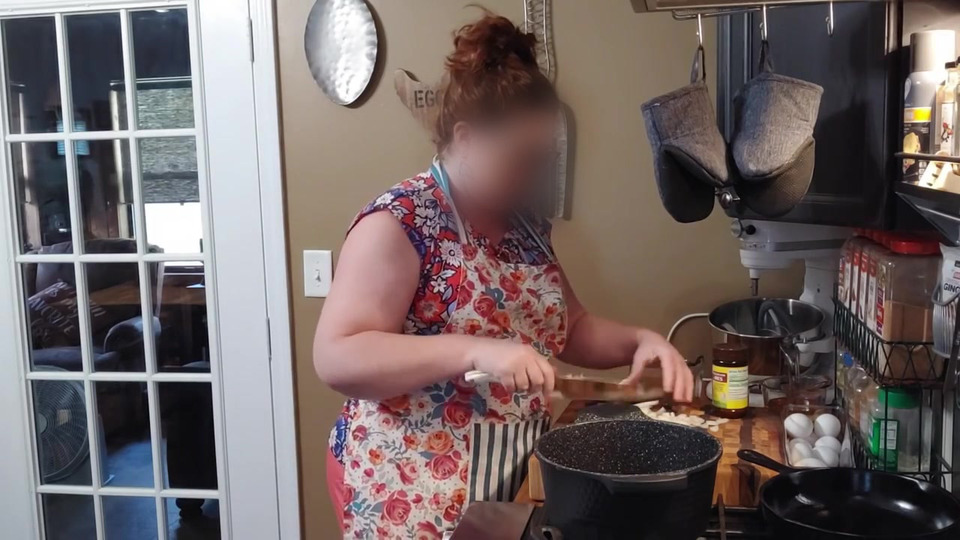} &
    \includegraphics[width=\linewidth,height=0.125\textheight,keepaspectratio]{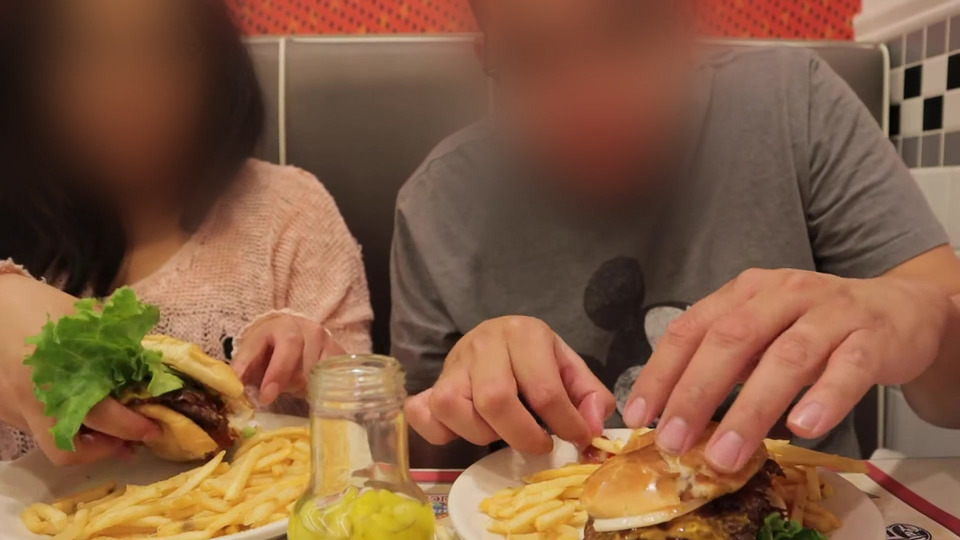} &
    \includegraphics[width=\linewidth,height=0.125\textheight,keepaspectratio]{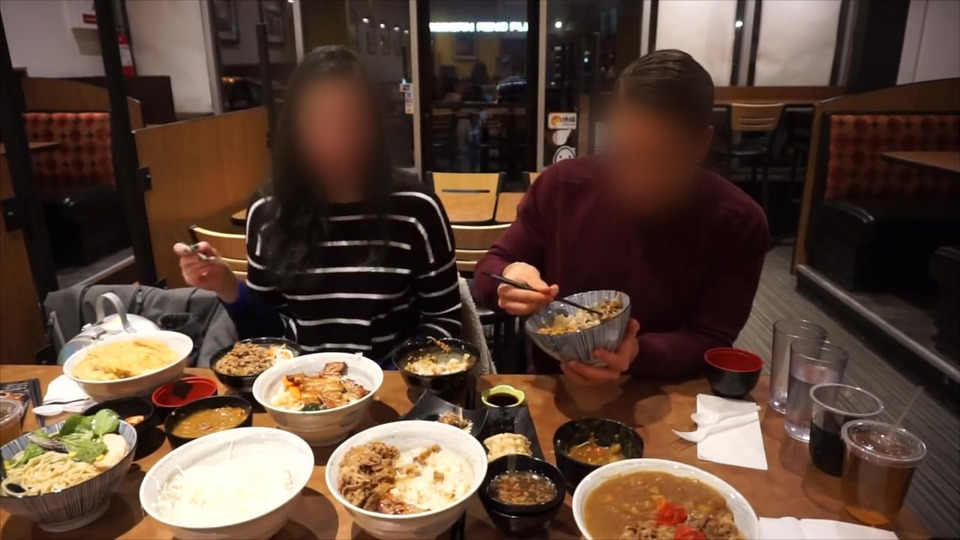} \\[2pt]

    \rotatebox{90}{2D Pose} &
    \includegraphics[width=\linewidth,height=0.125\textheight,keepaspectratio]{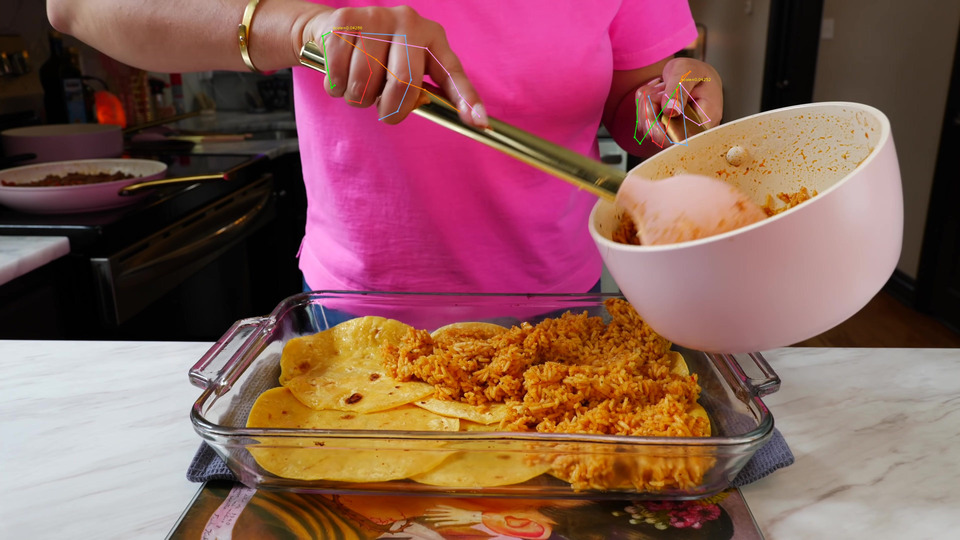} &
    \includegraphics[width=\linewidth,height=0.125\textheight,keepaspectratio]{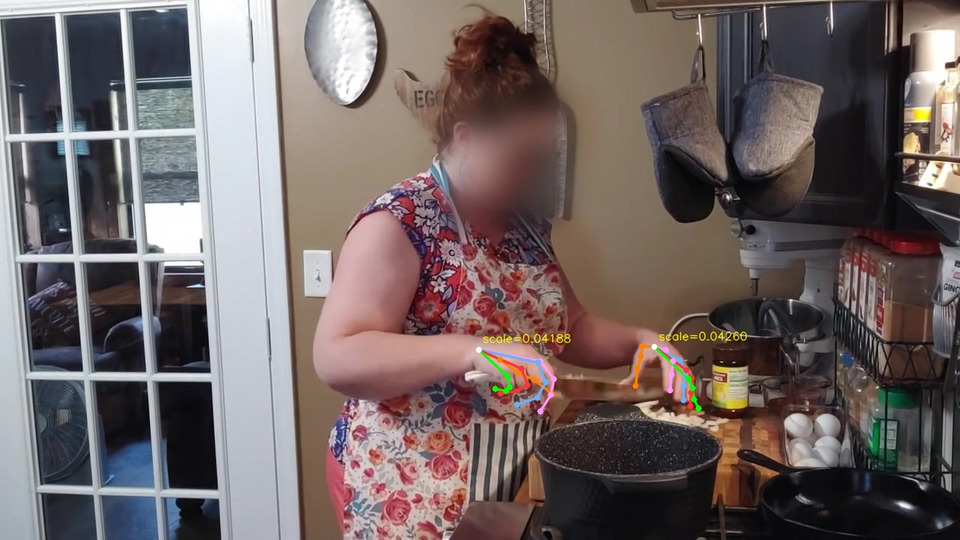} &
    \includegraphics[width=\linewidth,height=0.125\textheight,keepaspectratio]{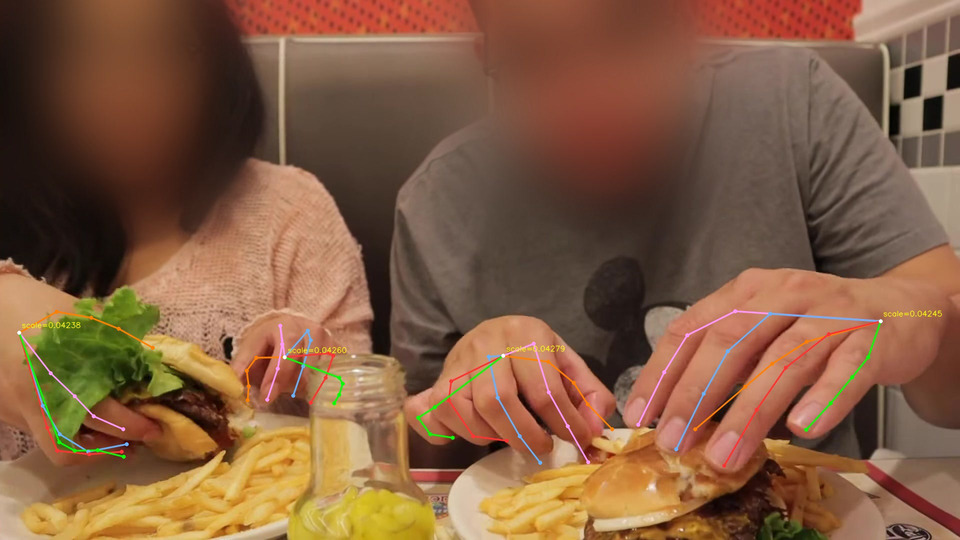} &
    \includegraphics[width=\linewidth,height=0.125\textheight,keepaspectratio]{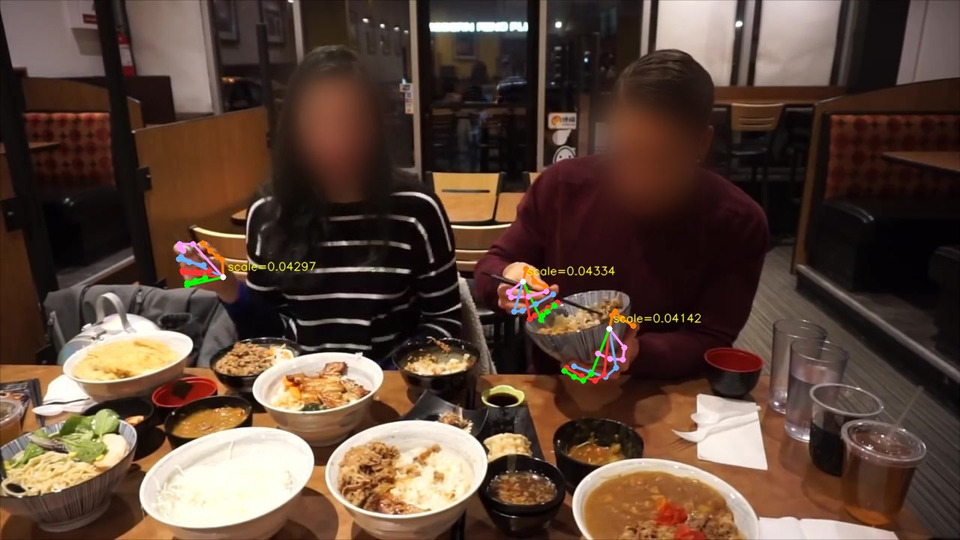} \\[2pt]

    \rotatebox{90}{3D Visualization} &
    \includegraphics[width=\linewidth,height=0.125\textheight,keepaspectratio]{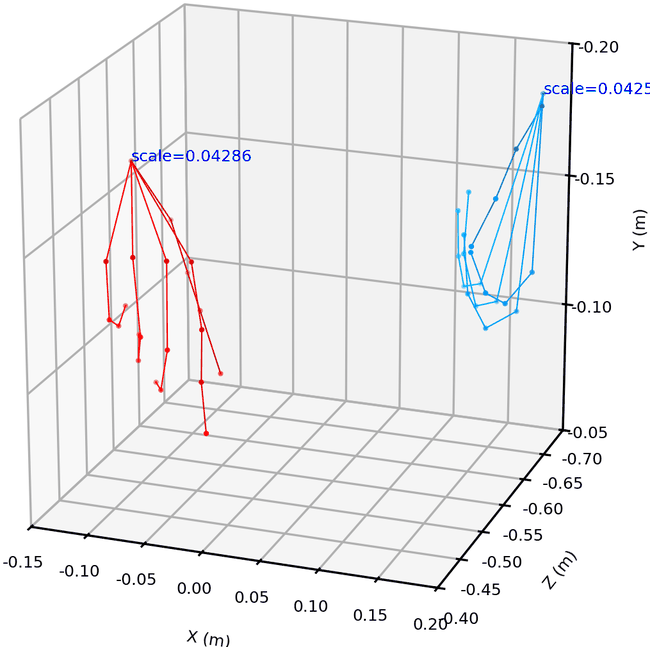} &
    \includegraphics[width=\linewidth,height=0.125\textheight,keepaspectratio]{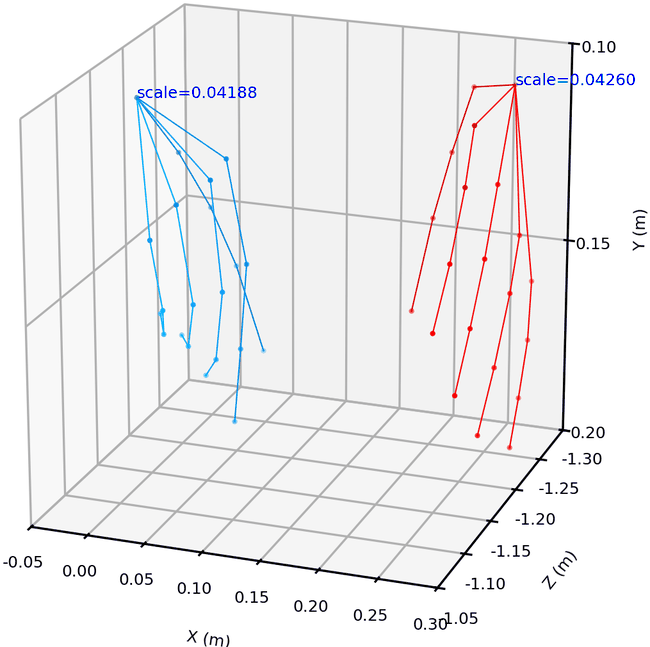} &
    \includegraphics[width=\linewidth,height=0.125\textheight,keepaspectratio]{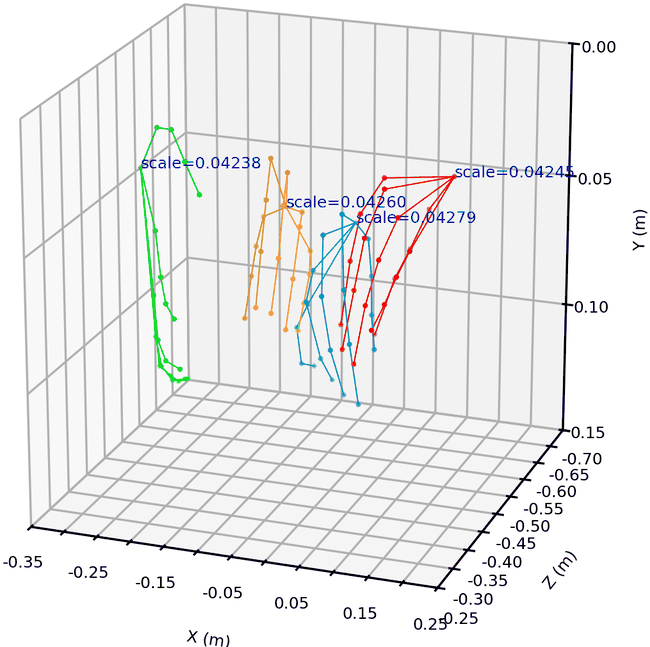} &
    \includegraphics[width=\linewidth,height=0.125\textheight,keepaspectratio]{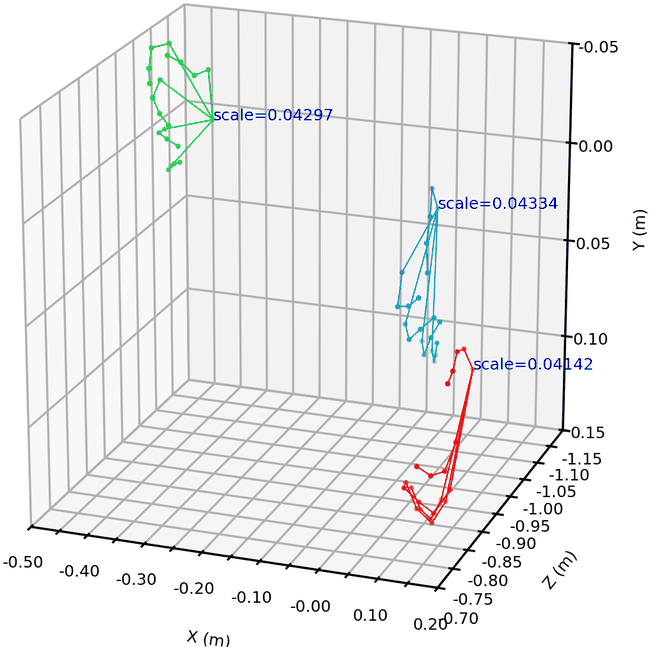} \\[2pt]

    \rotatebox{90}{Front View} &
    \includegraphics[width=\linewidth,height=0.125\textheight,keepaspectratio]{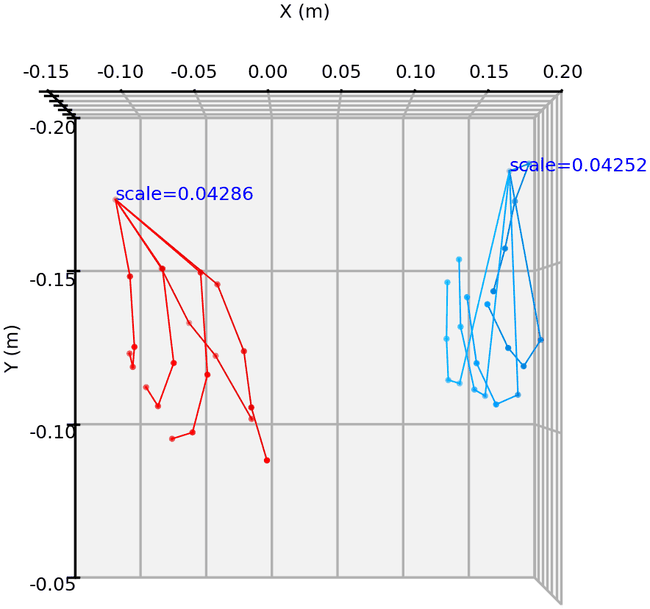} &
    \includegraphics[width=\linewidth,height=0.125\textheight,keepaspectratio]{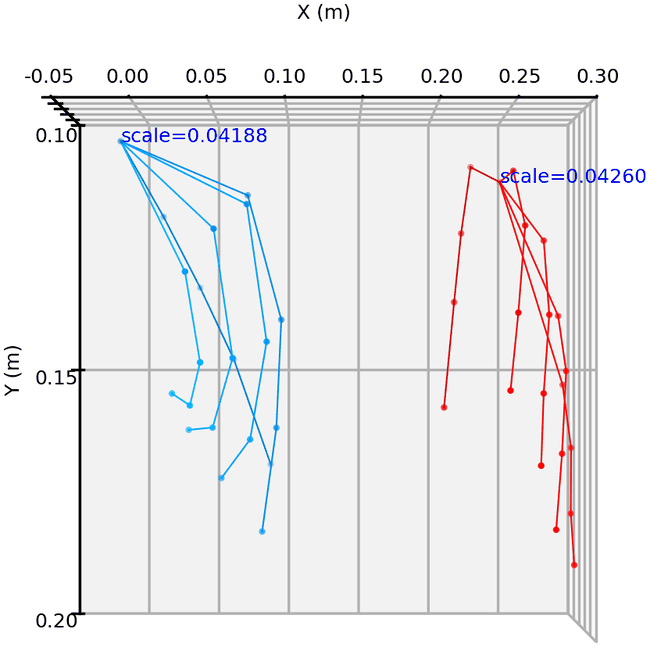} &
    \includegraphics[width=\linewidth,height=0.125\textheight,keepaspectratio]{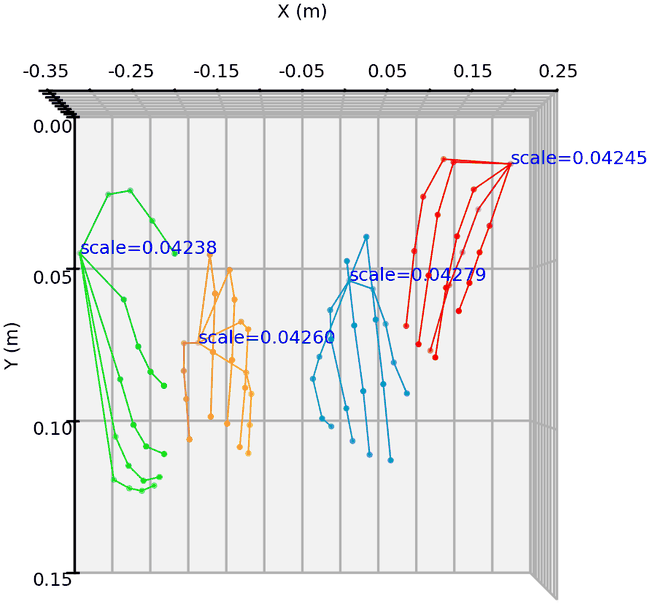} &
    \includegraphics[width=\linewidth,height=0.125\textheight,keepaspectratio]{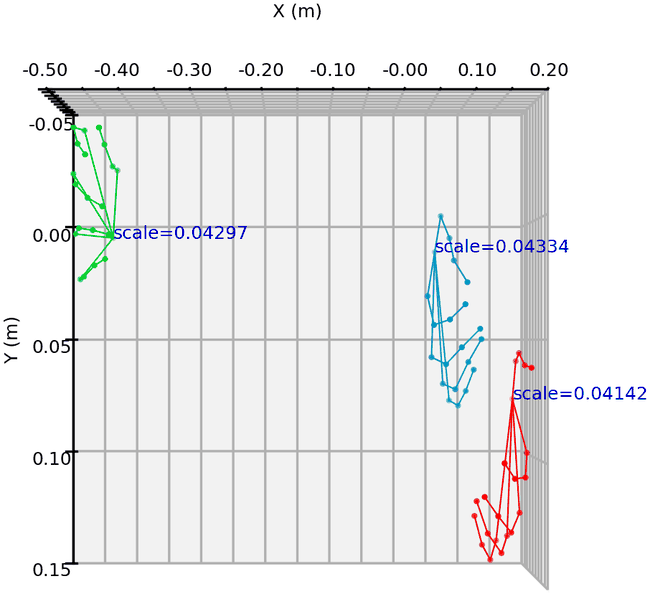} \\[2pt]

    \rotatebox{90}{Top View} &
    \includegraphics[width=\linewidth,height=0.125\textheight,keepaspectratio]{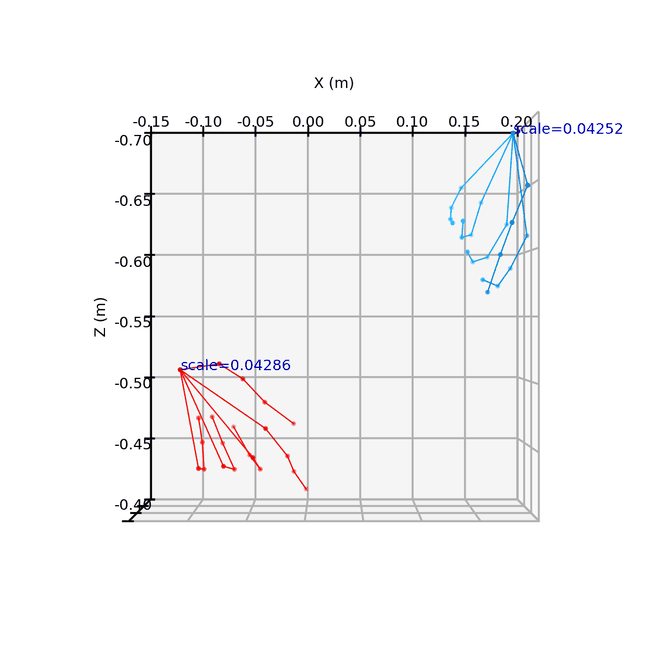} &
    \includegraphics[width=\linewidth,height=0.125\textheight,keepaspectratio]{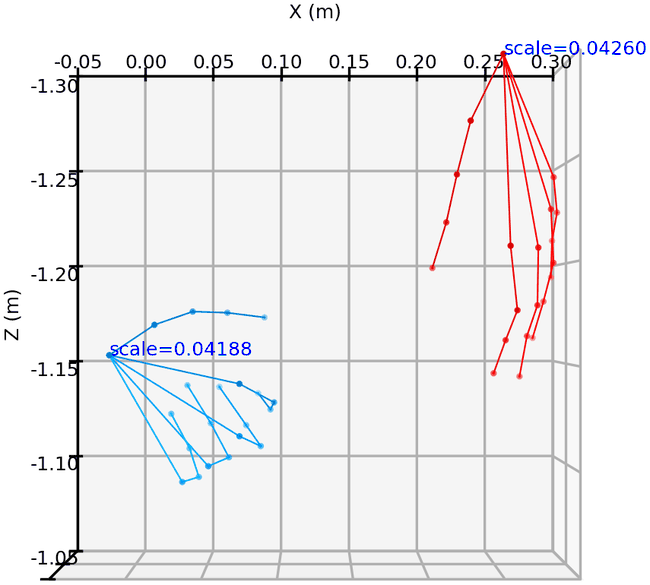} &
    \includegraphics[width=\linewidth,height=0.125\textheight,keepaspectratio]{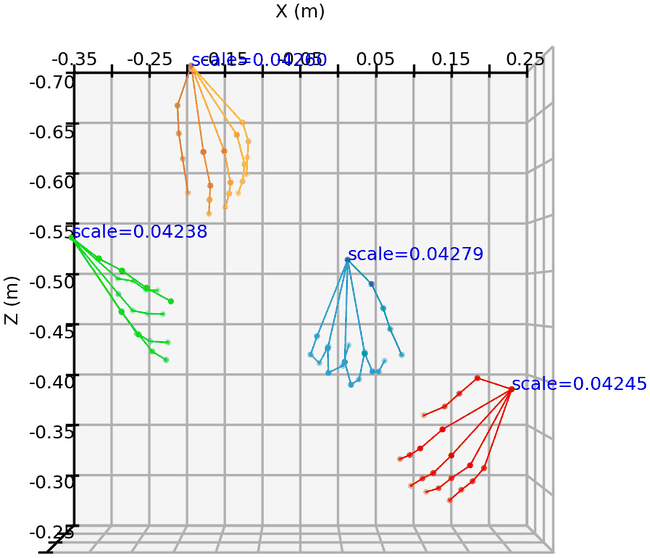} &
    \includegraphics[width=\linewidth,height=0.125\textheight,keepaspectratio]{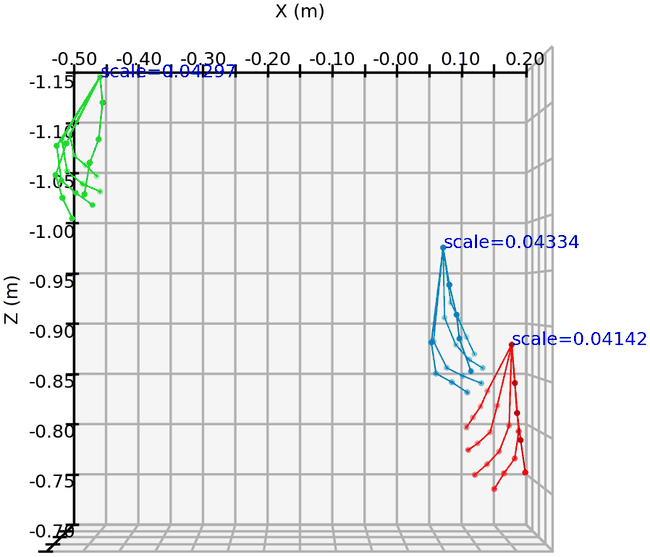} \\

    \end{tabular}

    \captionof{figure}{ScaleHP results on four in-the-wild examples (1/2).
    Best viewed zoomed in.}
    \label{fig:qualitative_results_1}
\end{center}

\clearpage
\begin{center}
    \centering
    \setlength{\tabcolsep}{1pt}
    \begin{tabular}{
        >{\centering\arraybackslash}m{0.025\textwidth}
        *{4}{>{\centering\arraybackslash}m{0.225\textwidth}}
    }

     & \textbf{Ex.5} & \textbf{Ex.6} & \textbf{Ex.7} & \textbf{Ex.8} \\[3pt]

    \rotatebox{90}{Input} &
    \includegraphics[width=\linewidth,height=0.125\textheight,keepaspectratio]{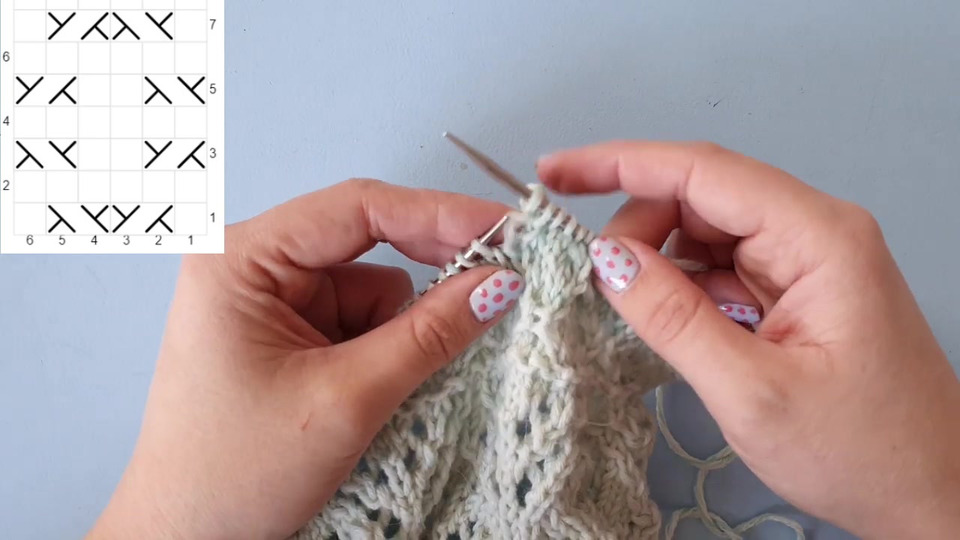} &
    \includegraphics[width=\linewidth,height=0.125\textheight,keepaspectratio]{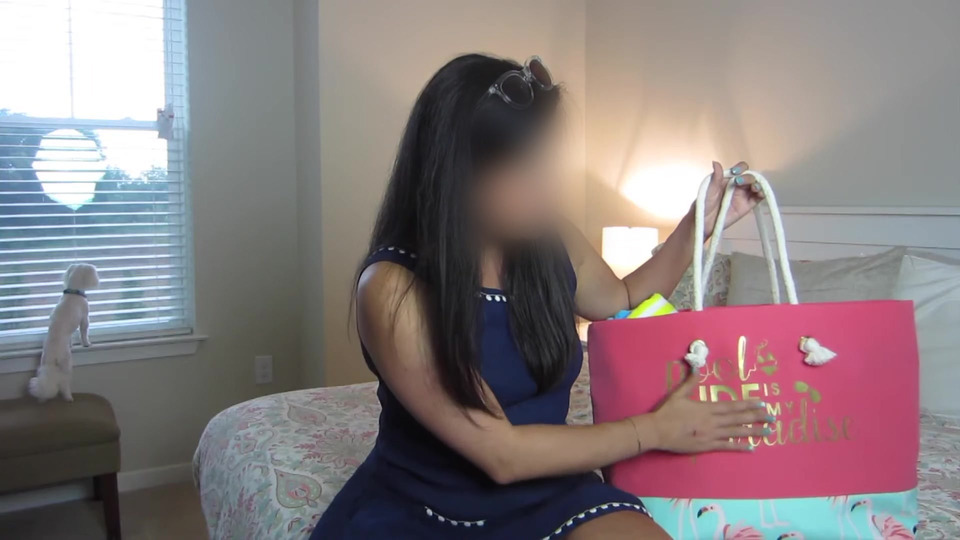} &
    \includegraphics[width=\linewidth,height=0.125\textheight,keepaspectratio]{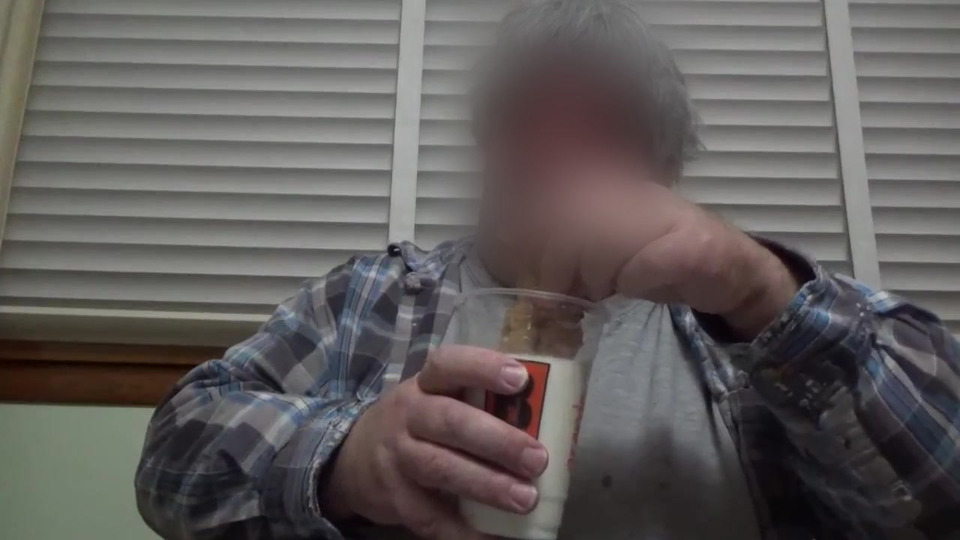} &
    \includegraphics[width=\linewidth,height=0.125\textheight,keepaspectratio]{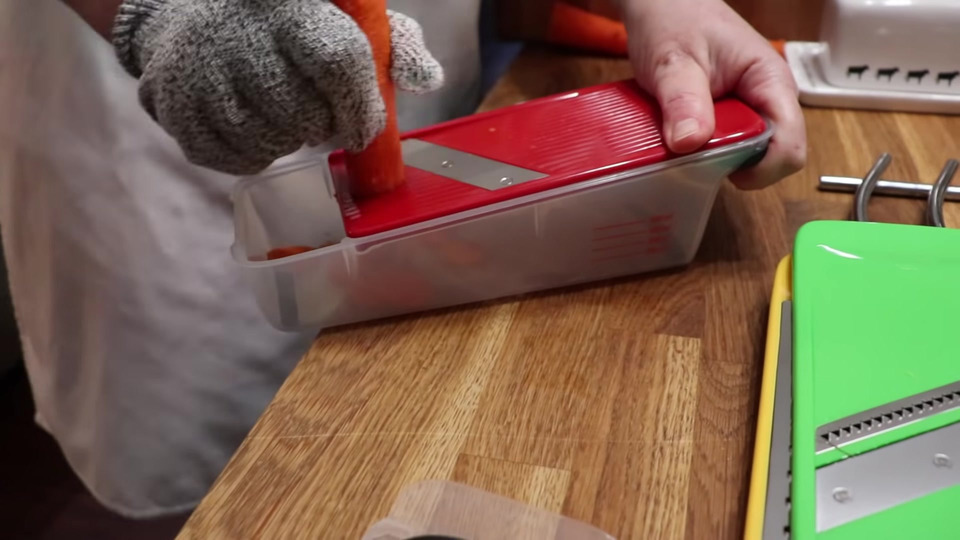} \\[2pt]

    \rotatebox{90}{2D Pose} &
    \includegraphics[width=\linewidth,height=0.125\textheight,keepaspectratio]{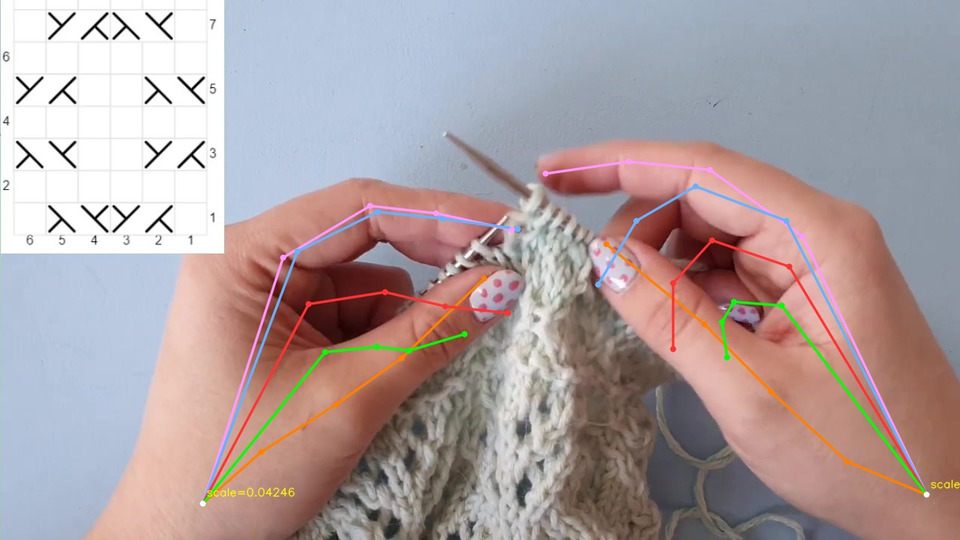} &
    \includegraphics[width=\linewidth,height=0.125\textheight,keepaspectratio]{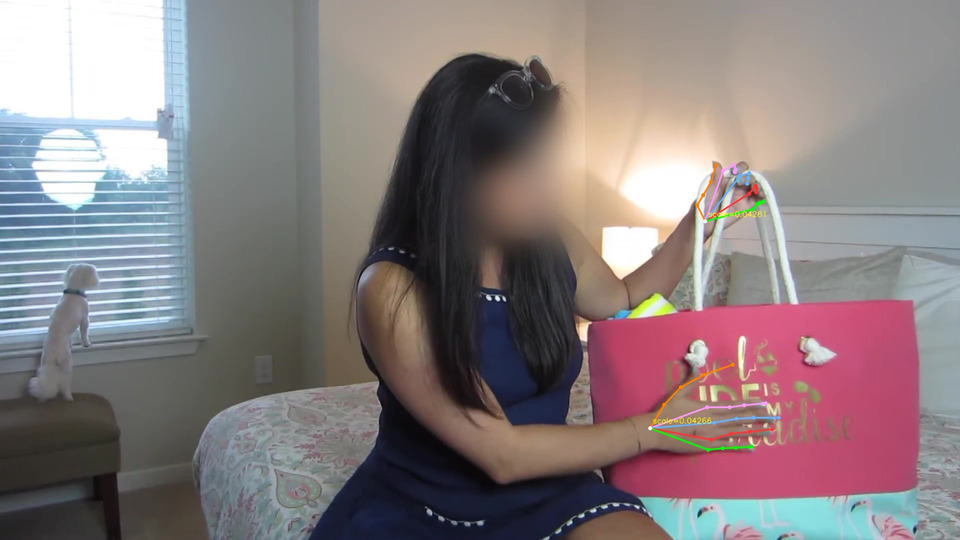} &
    \includegraphics[width=\linewidth,height=0.125\textheight,keepaspectratio]{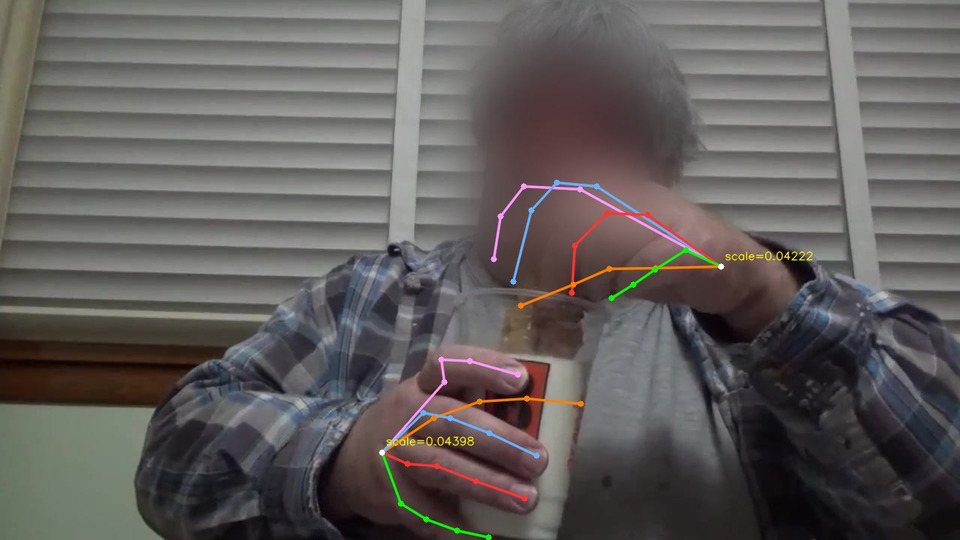} &
    \includegraphics[width=\linewidth,height=0.125\textheight,keepaspectratio]{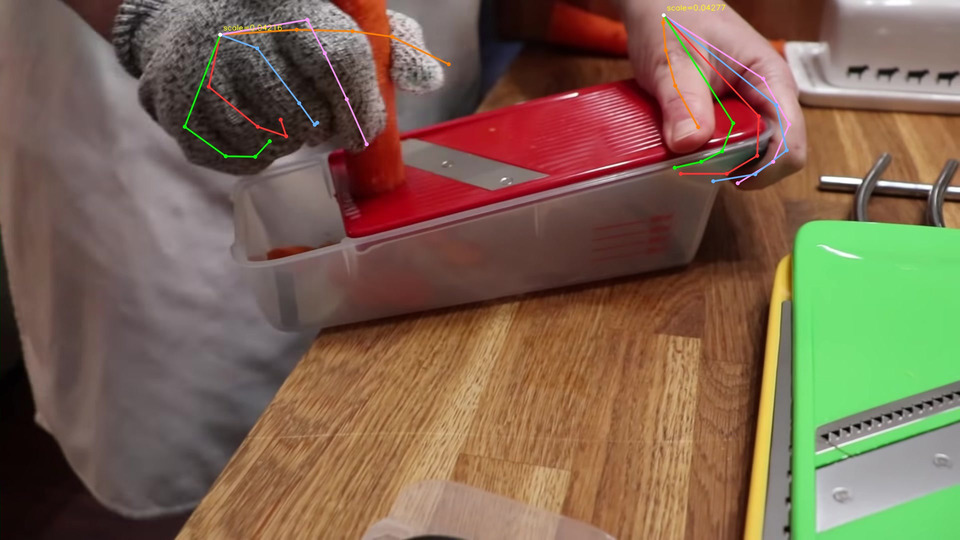} \\[2pt]

    \rotatebox{90}{3D Visualization} &
    \includegraphics[width=\linewidth,height=0.125\textheight,keepaspectratio]{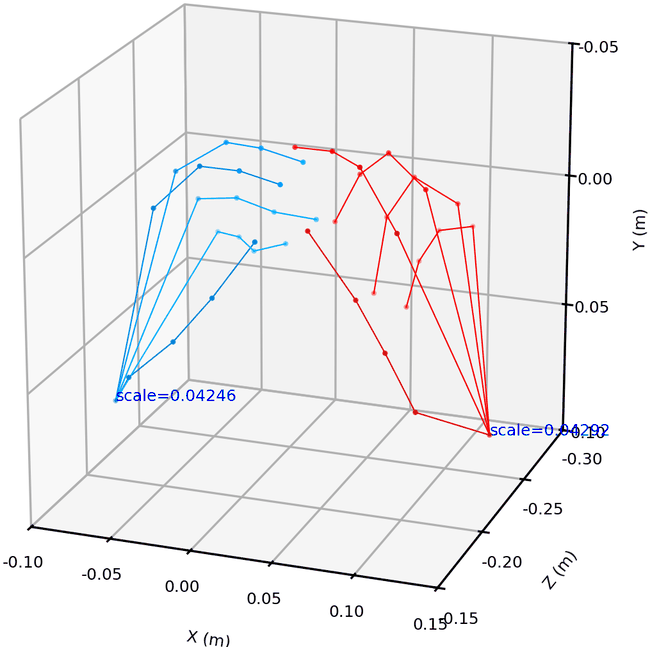} &
    \includegraphics[width=\linewidth,height=0.125\textheight,keepaspectratio]{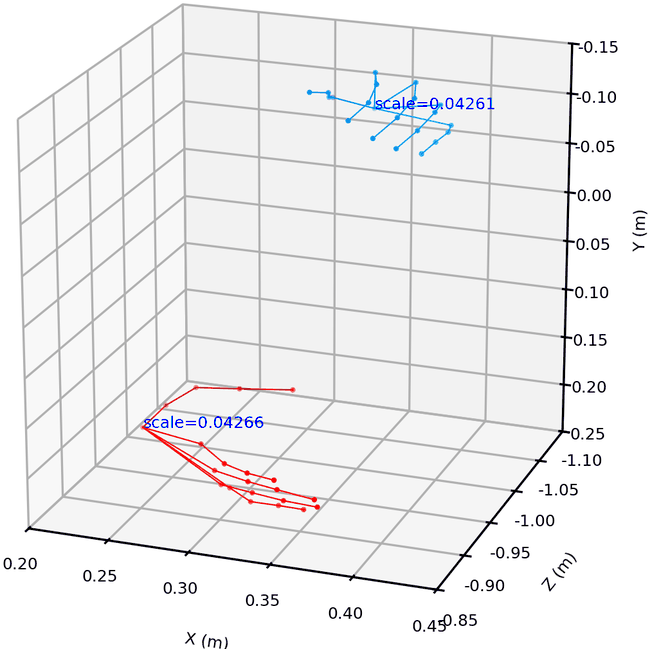} &
    \includegraphics[width=\linewidth,height=0.125\textheight,keepaspectratio]{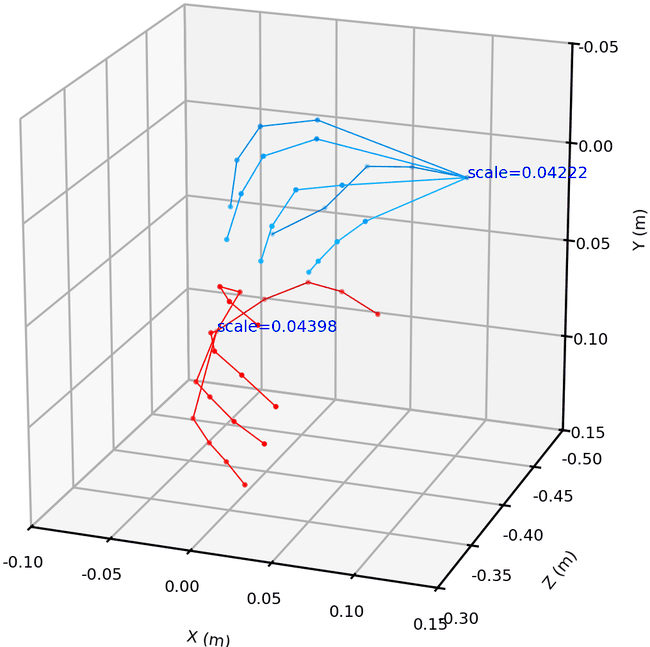} &
    \includegraphics[width=\linewidth,height=0.125\textheight,keepaspectratio]{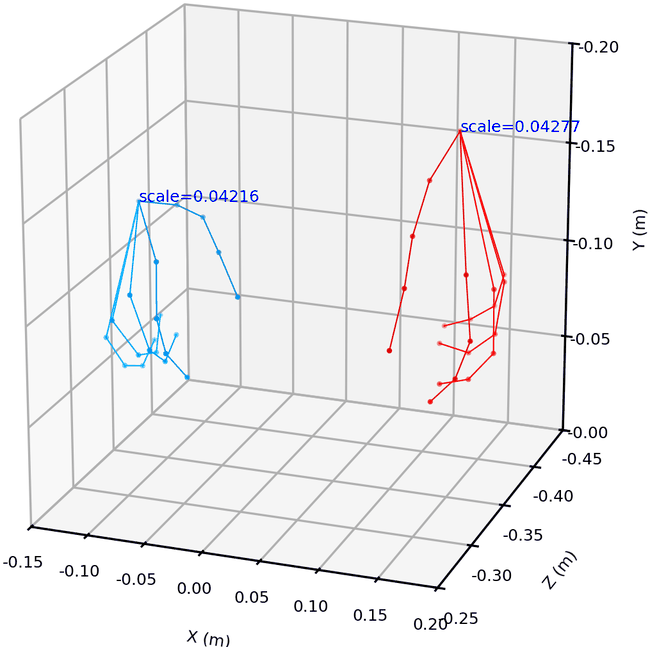} \\[2pt]

    \rotatebox{90}{Front View} &
    \includegraphics[width=\linewidth,height=0.125\textheight,keepaspectratio]{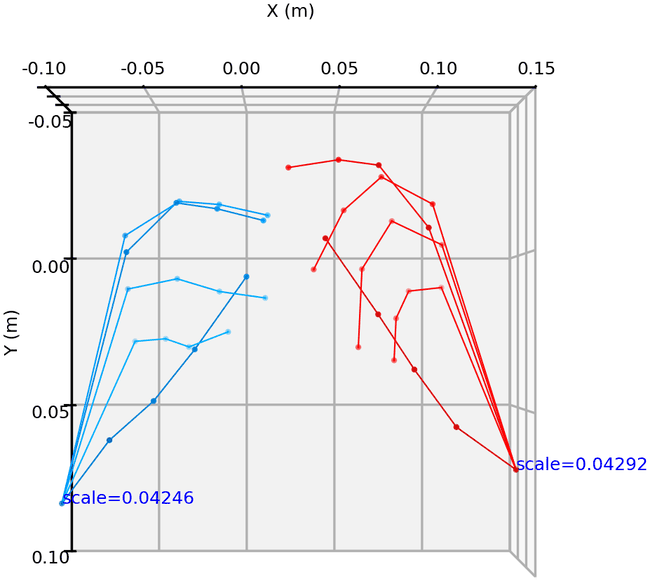} &
    \includegraphics[width=\linewidth,height=0.125\textheight,keepaspectratio]{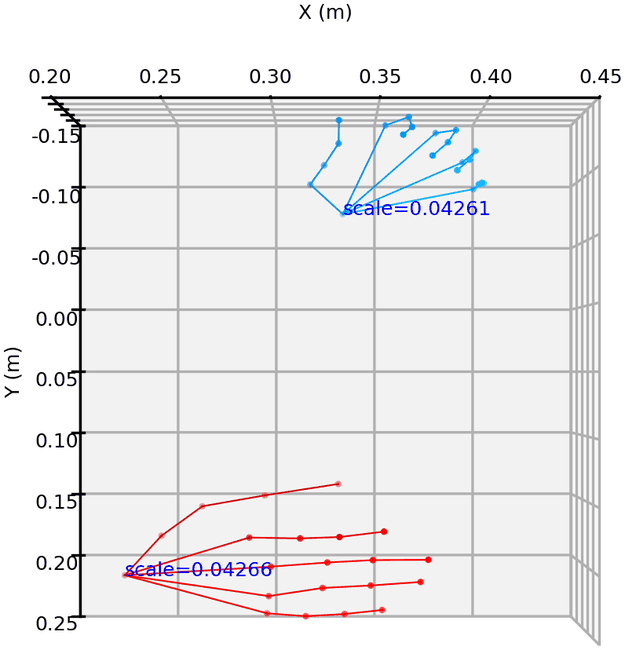} &
    \includegraphics[width=\linewidth,height=0.125\textheight,keepaspectratio]{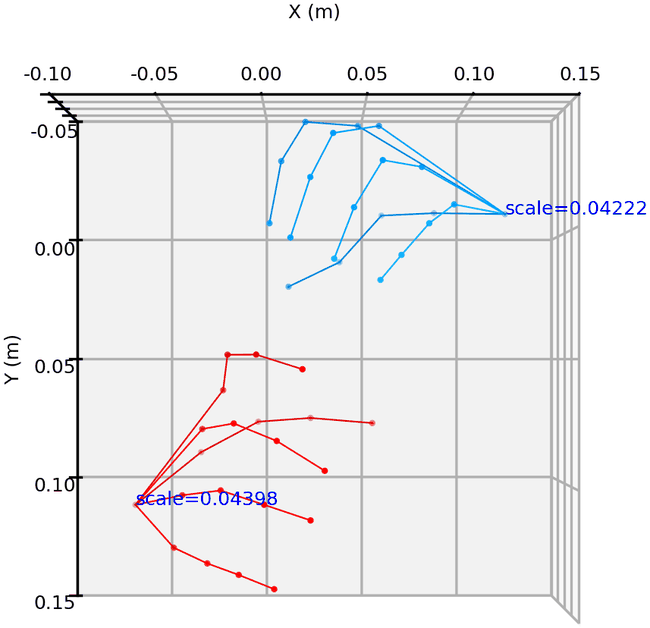} &
    \includegraphics[width=\linewidth,height=0.125\textheight,keepaspectratio]{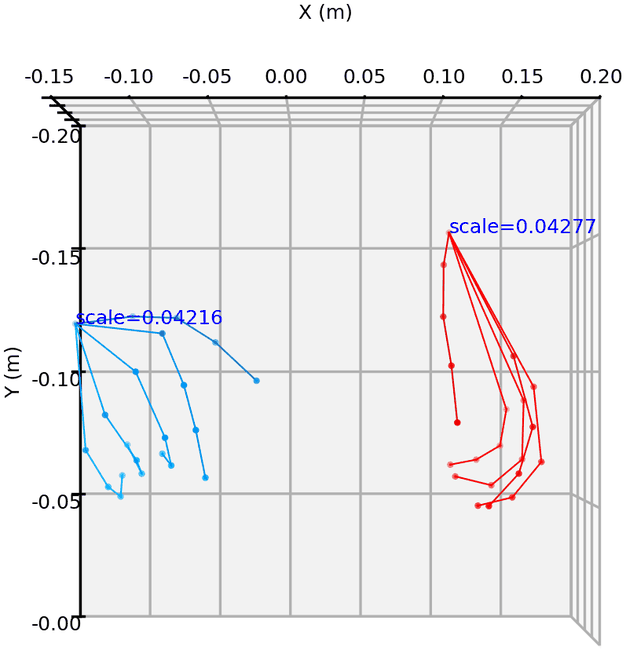} \\[2pt]

    \rotatebox{90}{Top View} &
    \includegraphics[width=\linewidth,height=0.125\textheight,keepaspectratio]{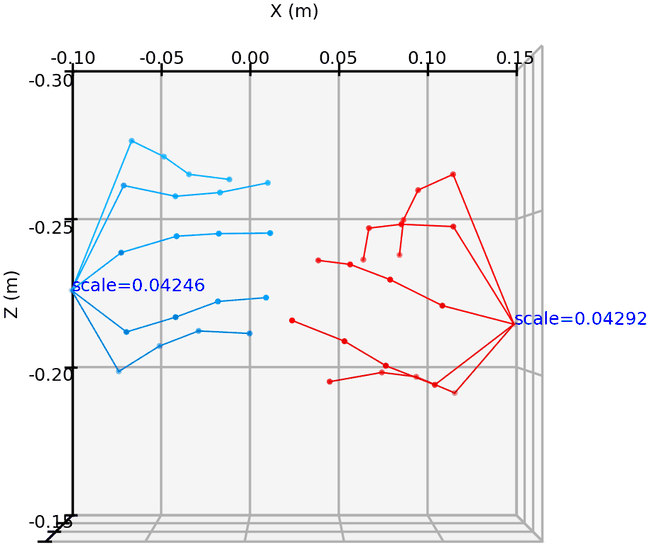} &
    \includegraphics[width=\linewidth,height=0.125\textheight,keepaspectratio]{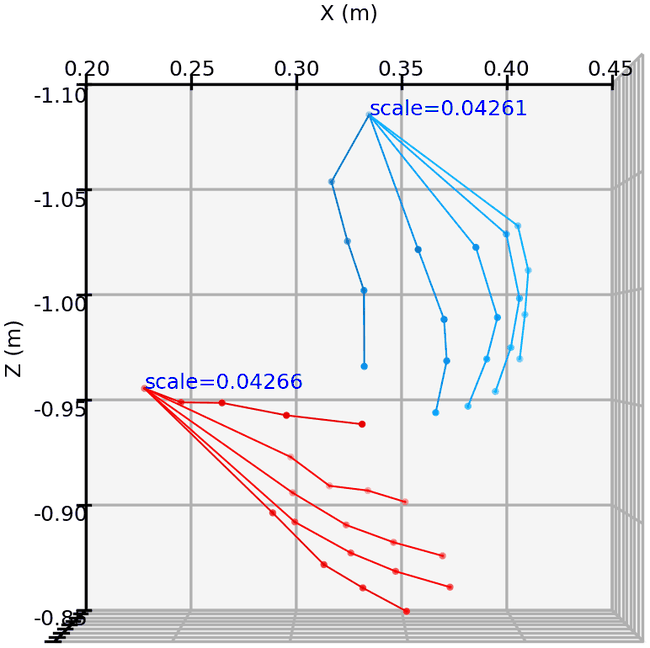} &
    \includegraphics[width=\linewidth,height=0.125\textheight,keepaspectratio]{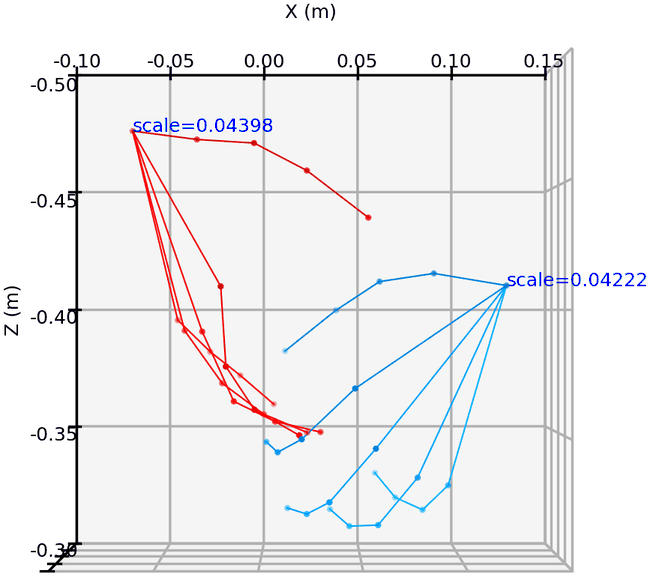} &
    \includegraphics[width=\linewidth,height=0.125\textheight,keepaspectratio]{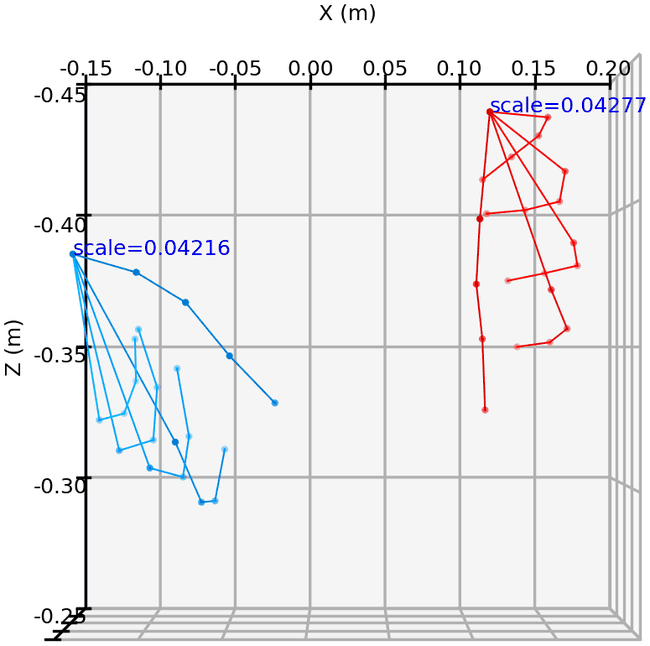} \\

    \end{tabular}

    \captionof{figure}{ScaleHP results on four in-the-wild examples (2/2).
    Best viewed zoomed in.}
    \label{fig:qualitative_results_2}
\end{center}

\clearpage
\twocolumn
\raggedbottom
\section{Limitations and Future Works}

\noindent\textbf{Limitations.}
Because the training datasets predominantly contain adult hands, metric-space
accuracy may degrade for infants, young children, and populations whose hand
size distribution is underrepresented during training. ScaleHP also assumes
known camera intrinsics, so focal-length or principal-point errors can propagate
to the recovered translation.

\noindent\textbf{Future Works.}
We plan to incorporate training data spanning broader ages and demographic
variation, evaluate robustness to camera-calibration perturbations, and extend
the frame-wise method with temporal constraints for stable video tracking.